%% file: TRO_final_ArXiv.tex
\documentclass[journal]{IEEEtran}

\usepackage{hyperref}
\usepackage{graphicx}
\usepackage{amsmath}
\usepackage{color}
\include{mathsym}

\usepackage{enumerate}   

\newtheorem{propertynew}{Property}
\newtheorem{proposition}{Proposition}

\def\bull{\vrule height .9ex width .8ex depth -.1ex} 

\newenvironment{braced}
 {\par\smallskip\hbox to\columnwidth\bgroup
  \hss$\left\{\begin{minipage}{\columnwidth}}
 {\end{minipage}\right.$\hss\egroup\smallskip}

\def\BlockSchemeDoubleStage{\centering\includegraphics[width=\textwidth]{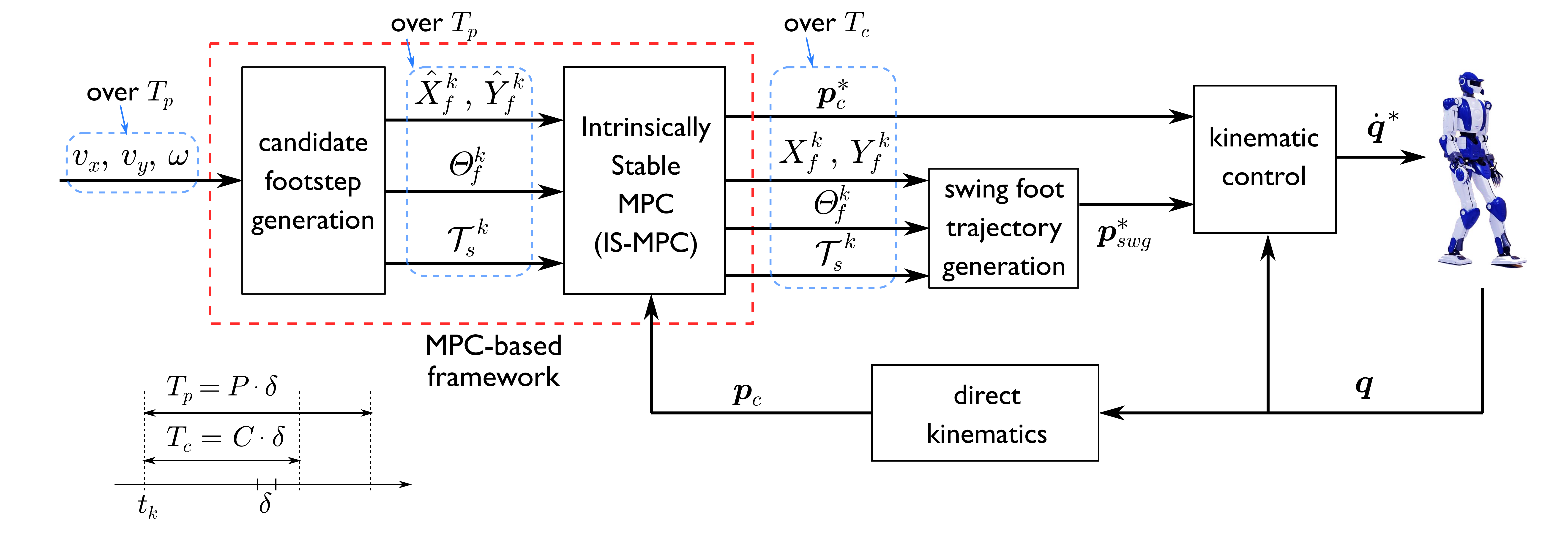}}
\def\LIPM_robot{\centering\includegraphics[width=0.3\textwidth]{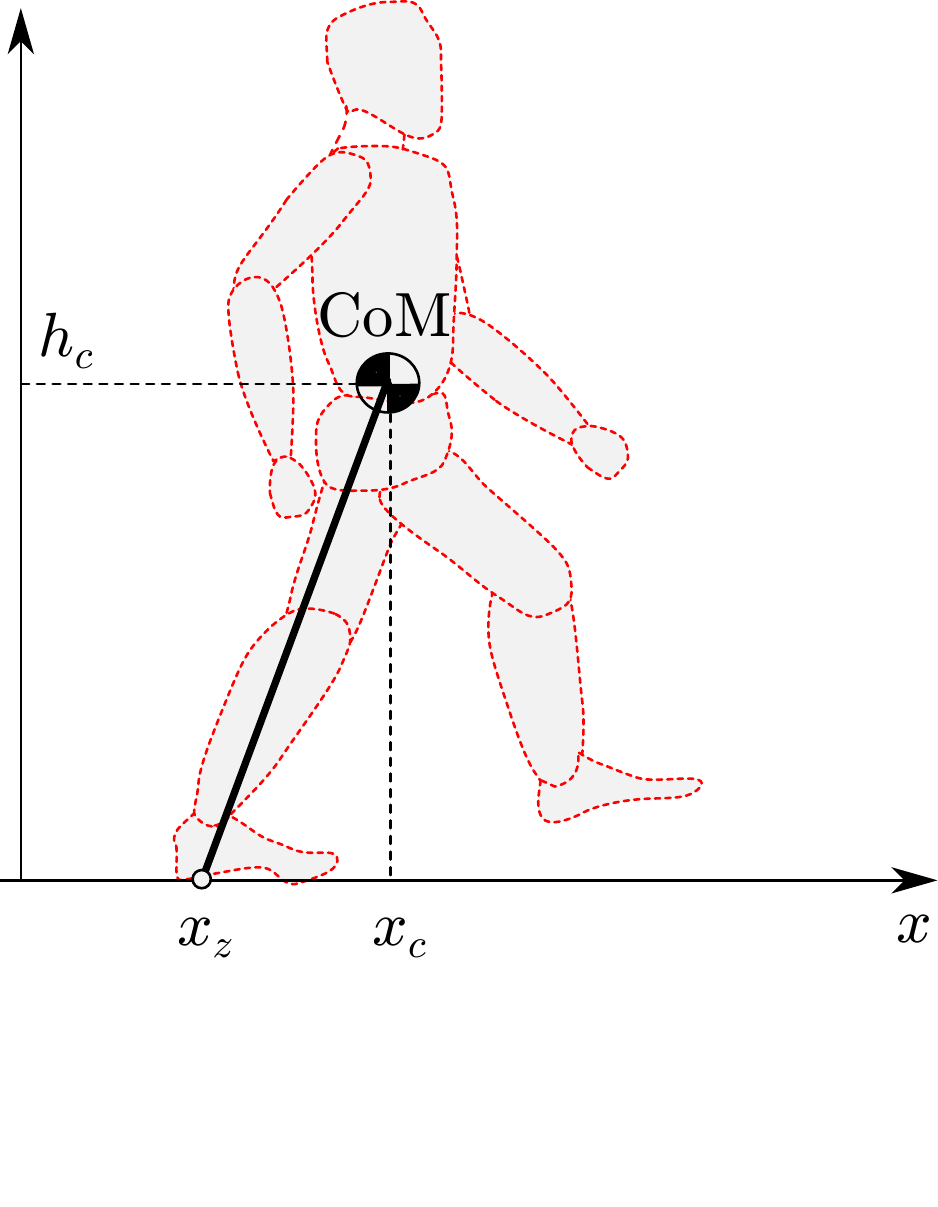}}
\def\PiecewiseConstZMPVel{\centering\includegraphics[width=\columnwidth]{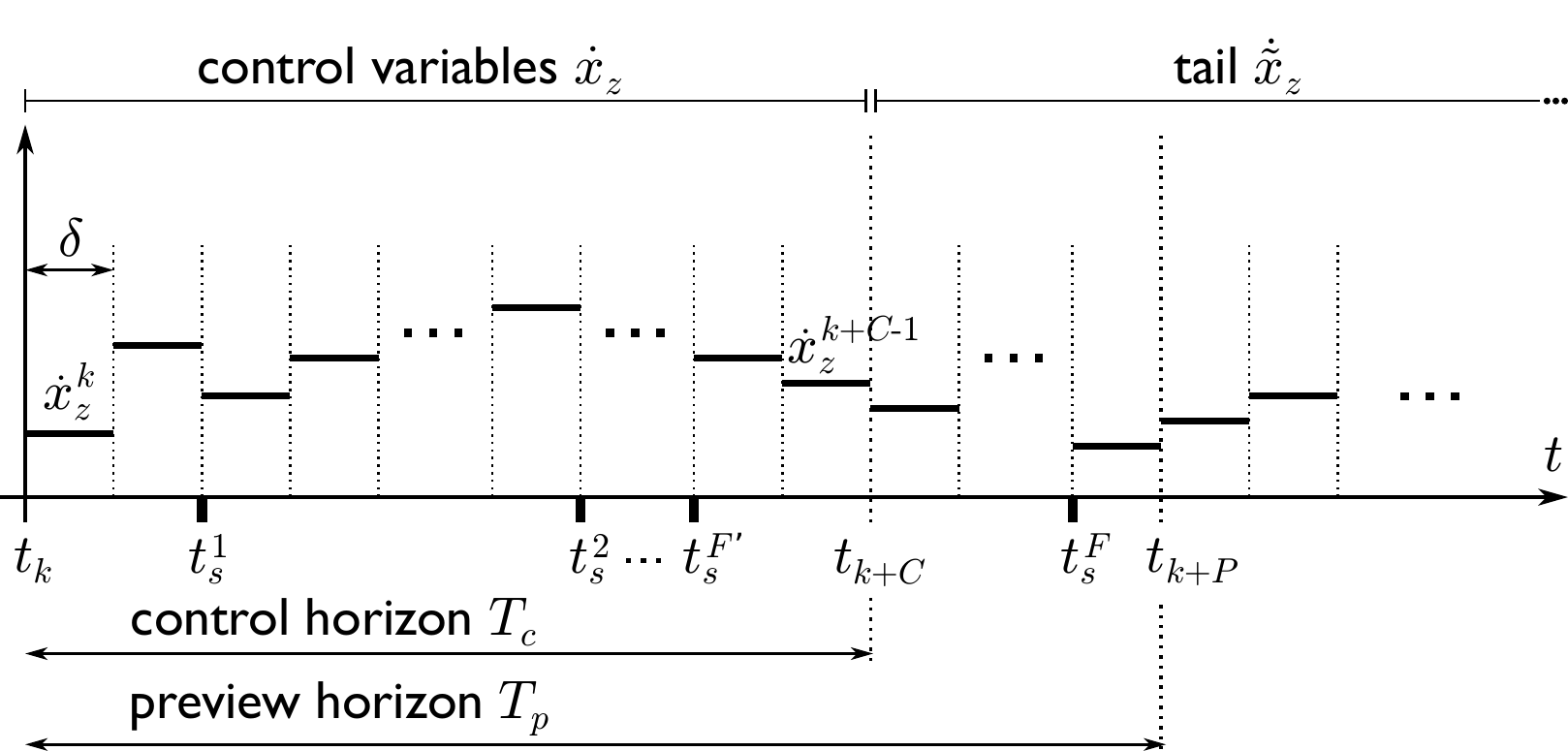}}
\def\DoubleSupport{\centering\includegraphics[width=0.8\columnwidth]{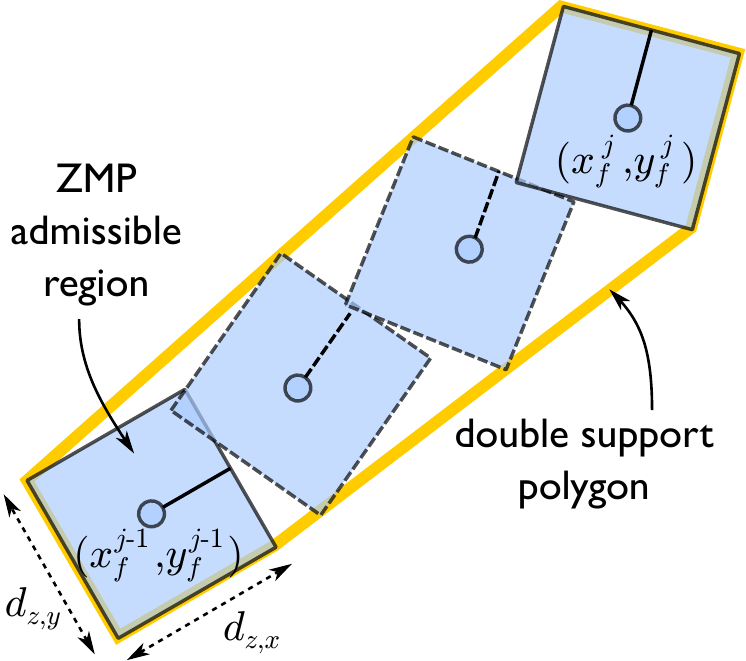}}
\def\KinematicConstraint{\centering\includegraphics[width=0.8\columnwidth]{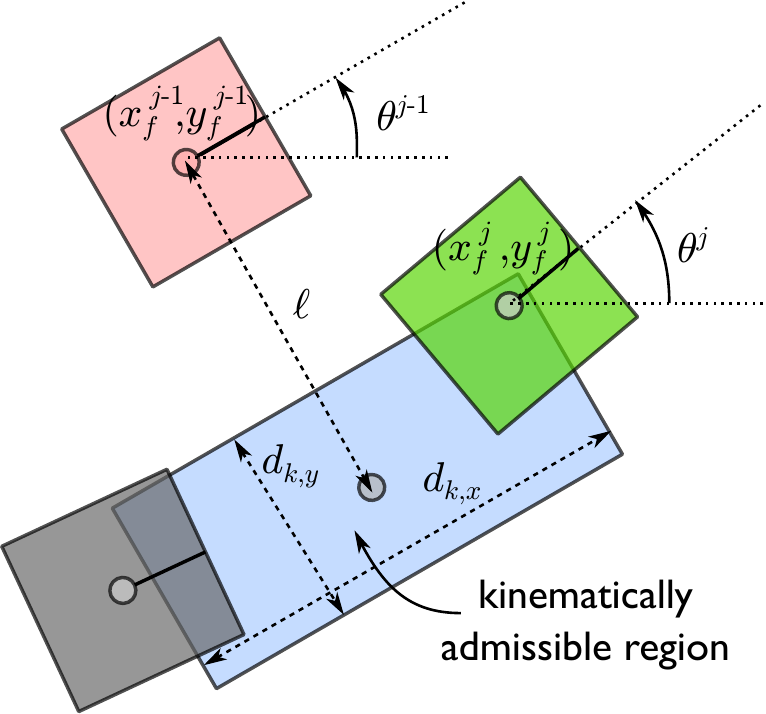}}
\def\stableLong{\centering\includegraphics[width=\columnwidth]{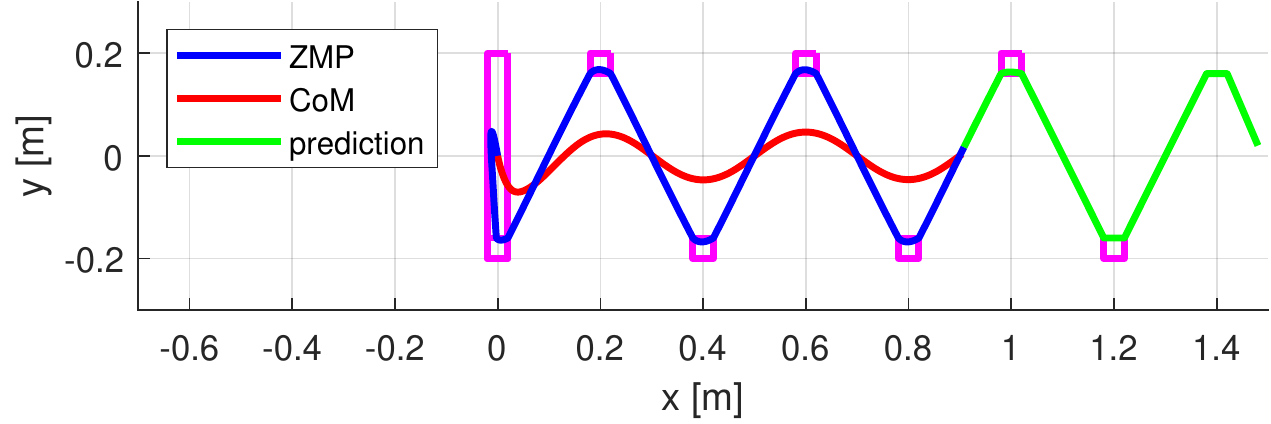}}
\def\stableShort{\centering\includegraphics[width=\columnwidth]{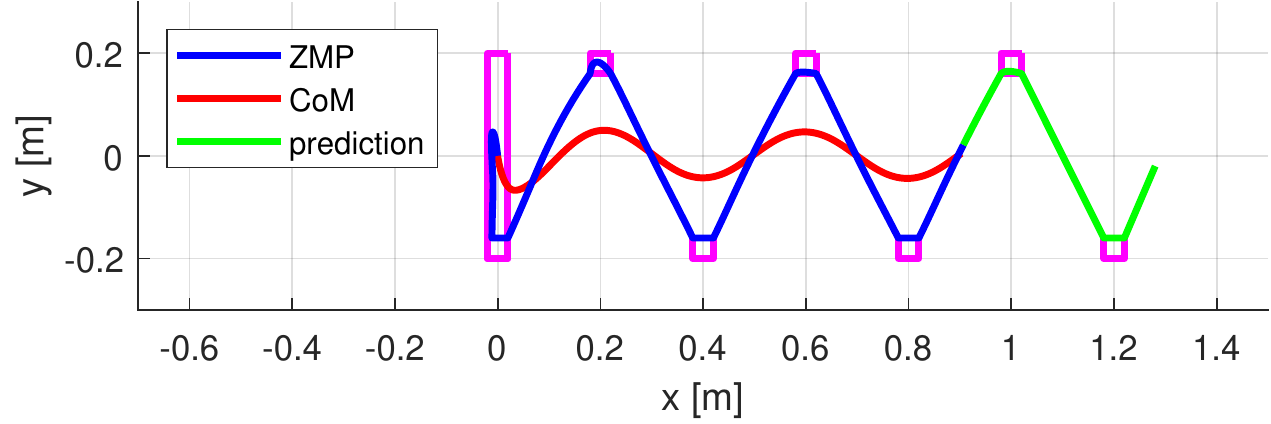}}
\def\jerkLong{\centering\includegraphics[width=\columnwidth]{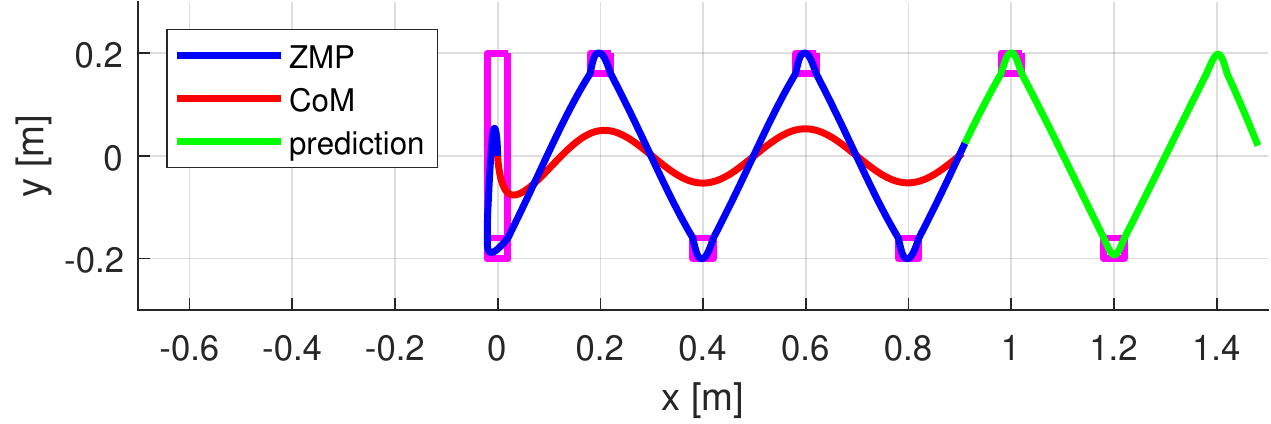}}
\def\jerkShort{\centering\includegraphics[width=\columnwidth]{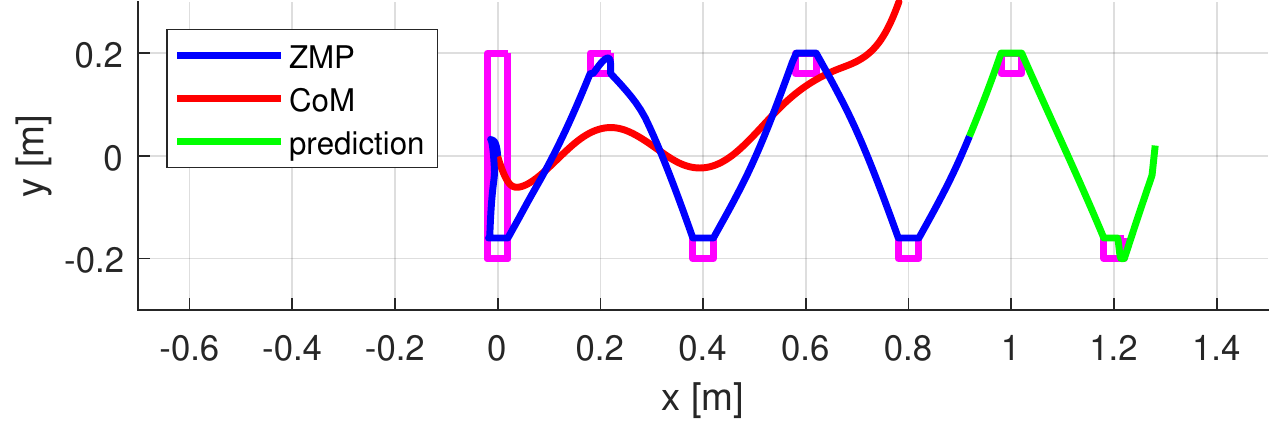}}
\def\jerkLongHighCoM{\centering\includegraphics[width=\columnwidth]{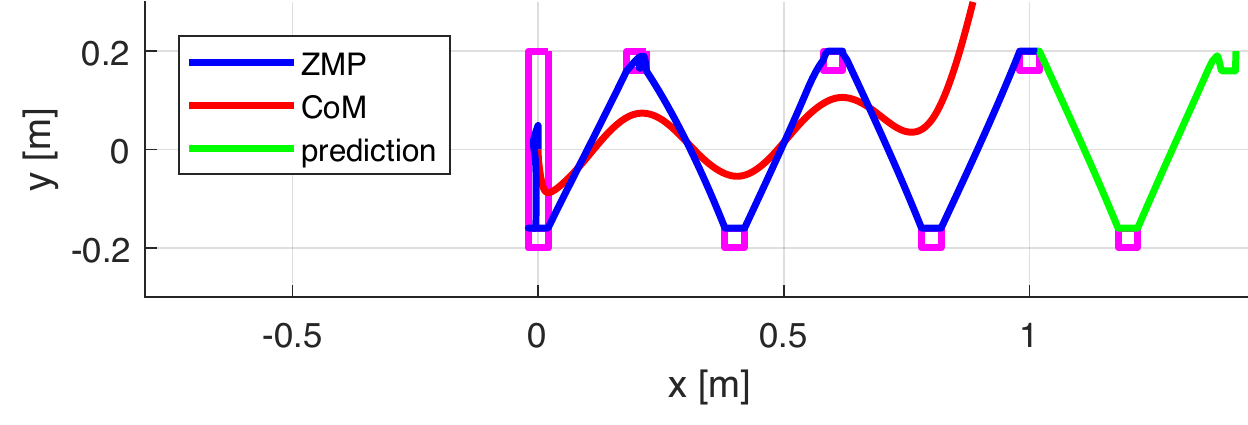}}
\def\stableLongHighCoM{\centering\includegraphics[width=\columnwidth]{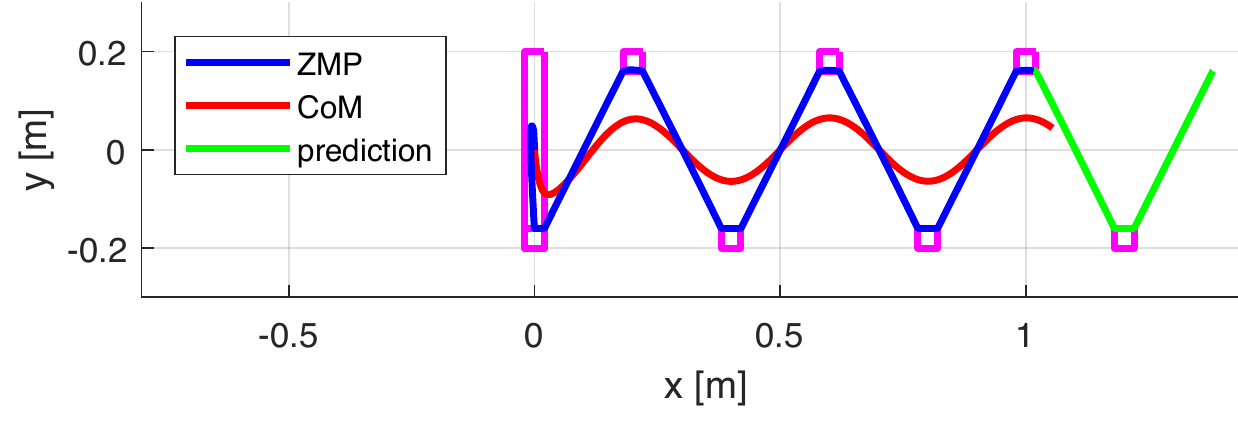}}
\def\footstepTimingRule{\centering\includegraphics[width=\columnwidth]{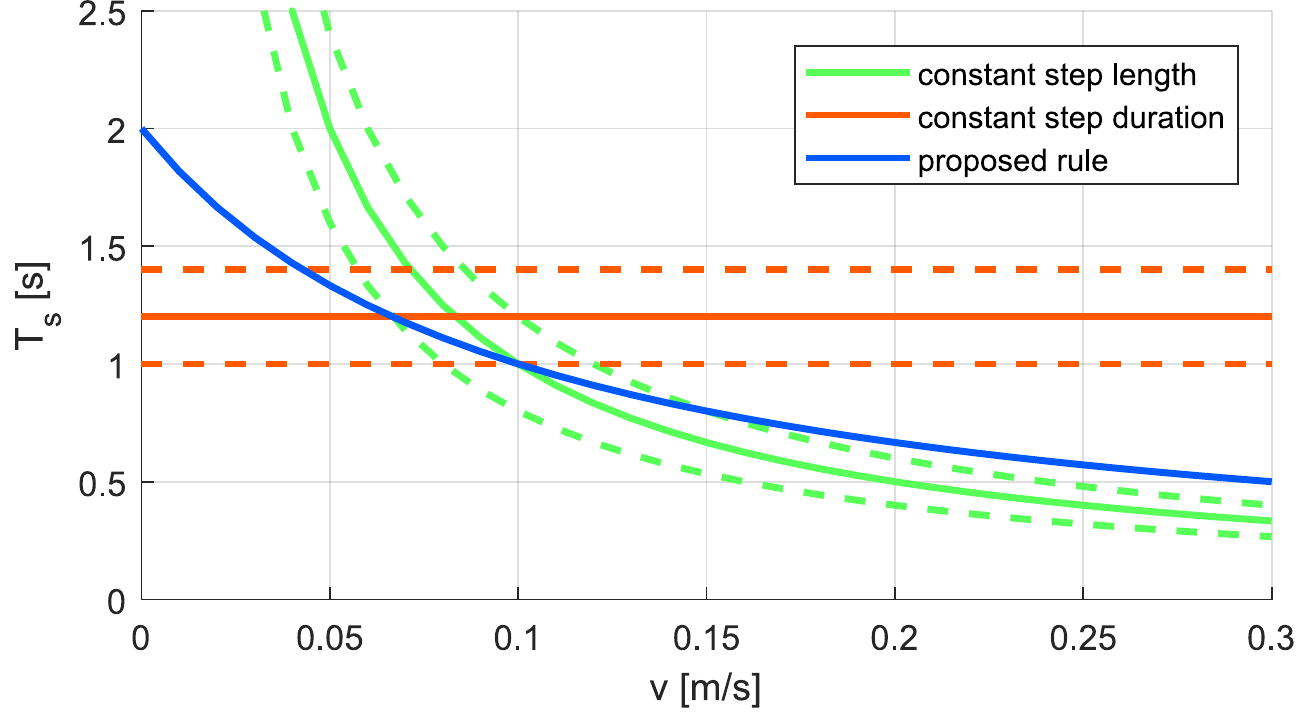}}
\def\stableSimKeepZMP{\centering\includegraphics[width=\columnwidth]{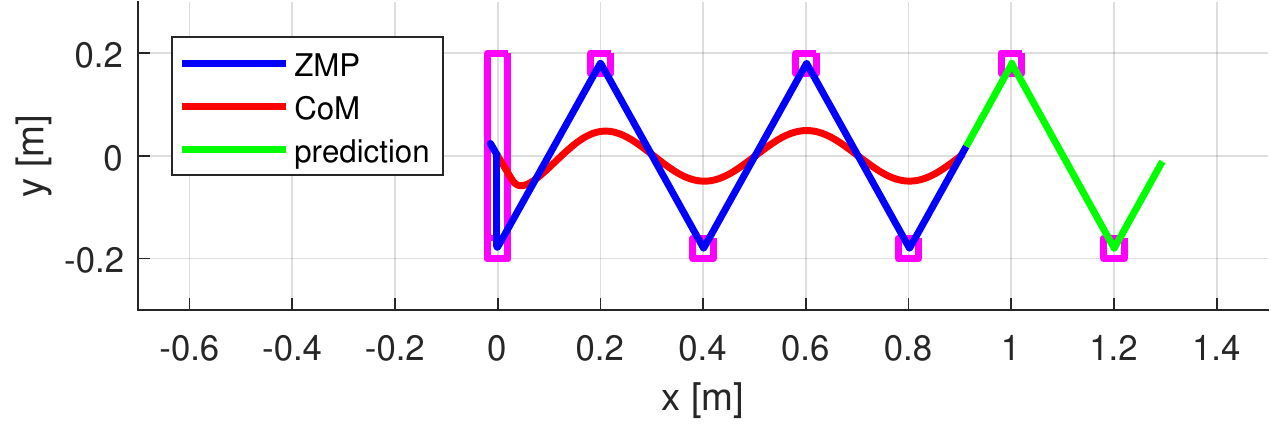}}
\def\jerkSimKeepZMP{\centering\includegraphics[width=\columnwidth]{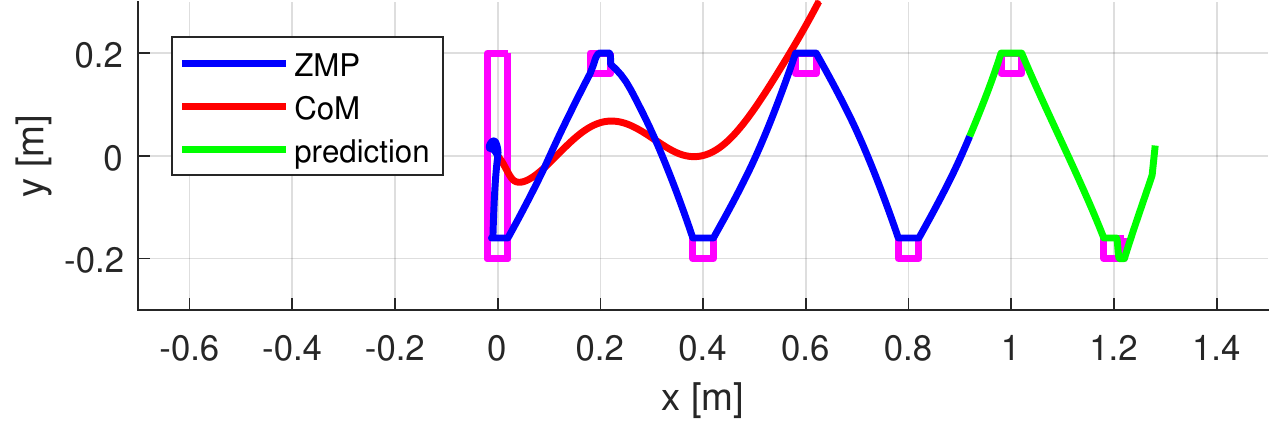}}

\def\FeasibilityRegions{\centering\includegraphics[width=\columnwidth]{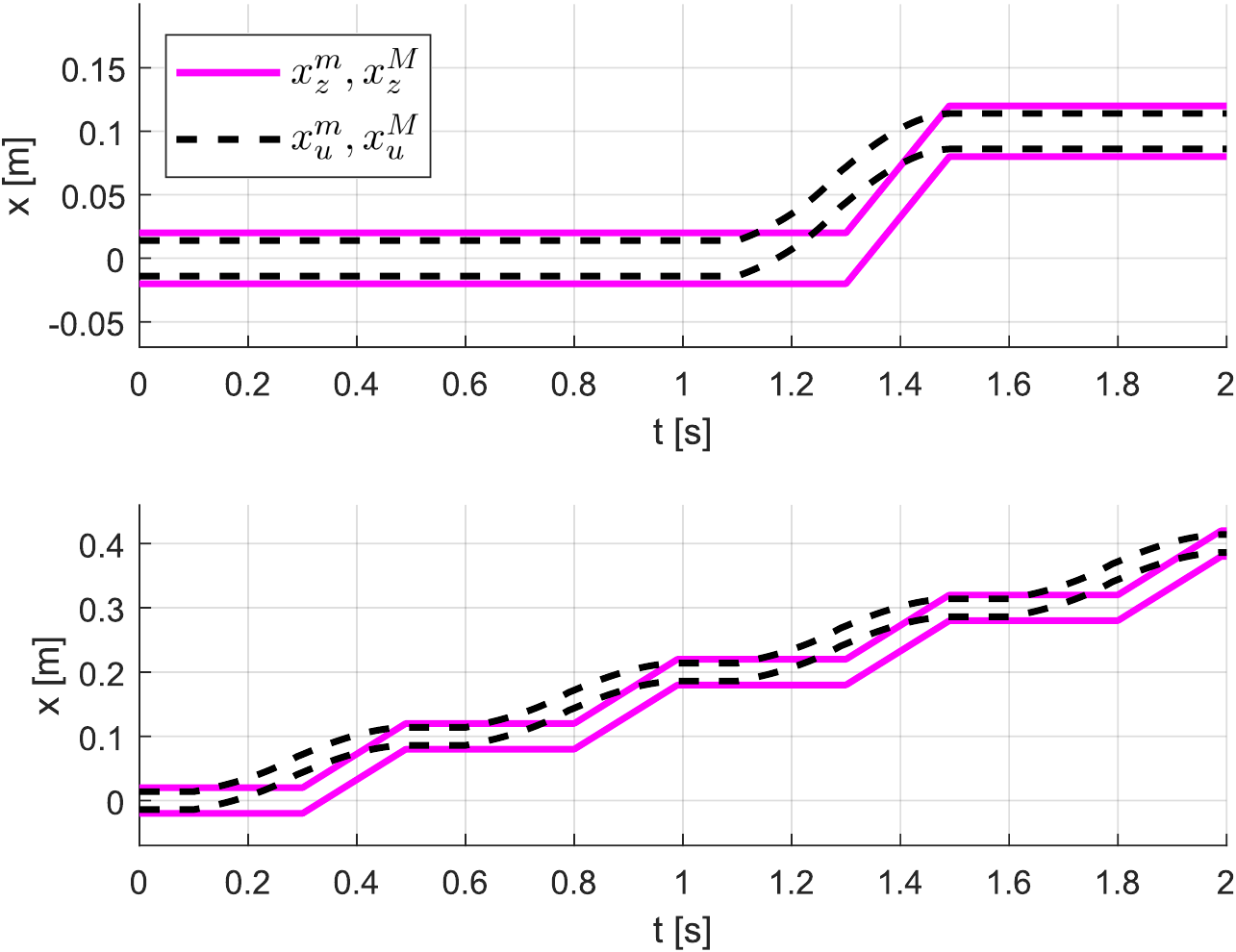}}

\def\fsGenCurve{\centering\includegraphics[width=0.8\columnwidth]{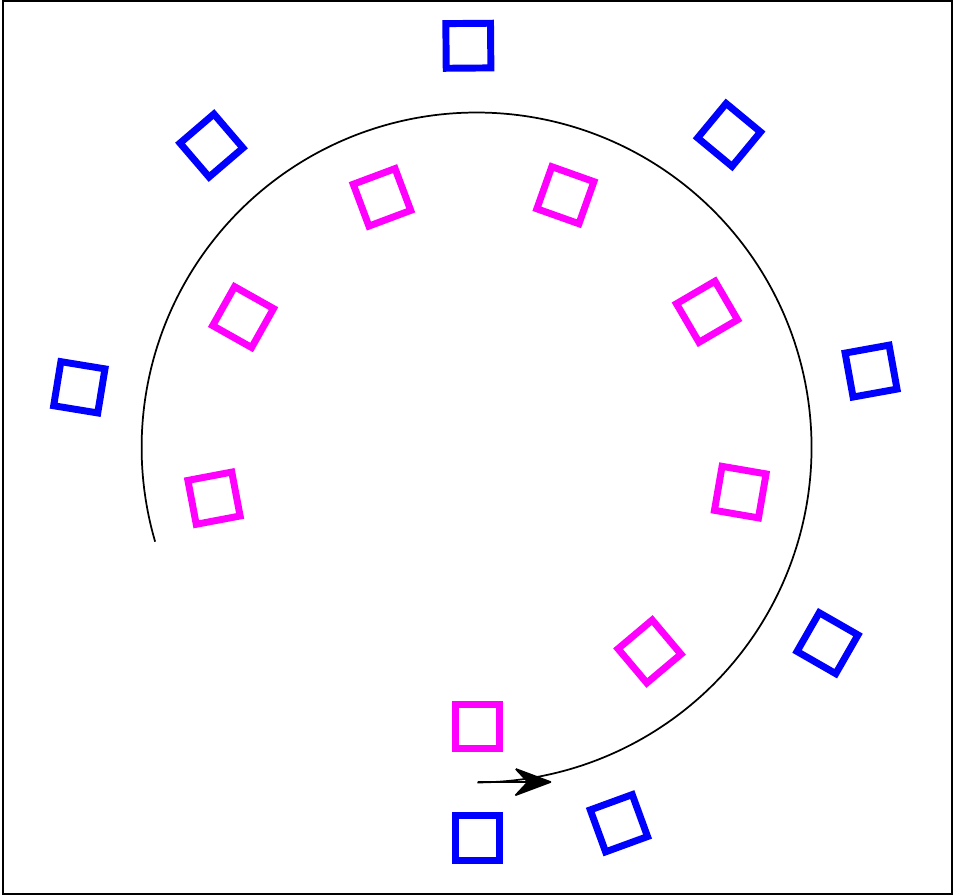}}
\def\fsGenLshape{\centering\includegraphics[width=0.8\columnwidth]{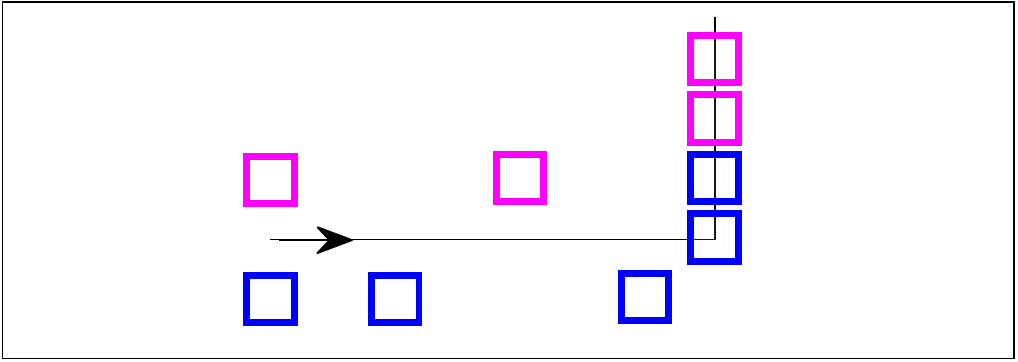}}
\def\fsGenDiagonal{\centering\includegraphics[width=0.8\columnwidth]{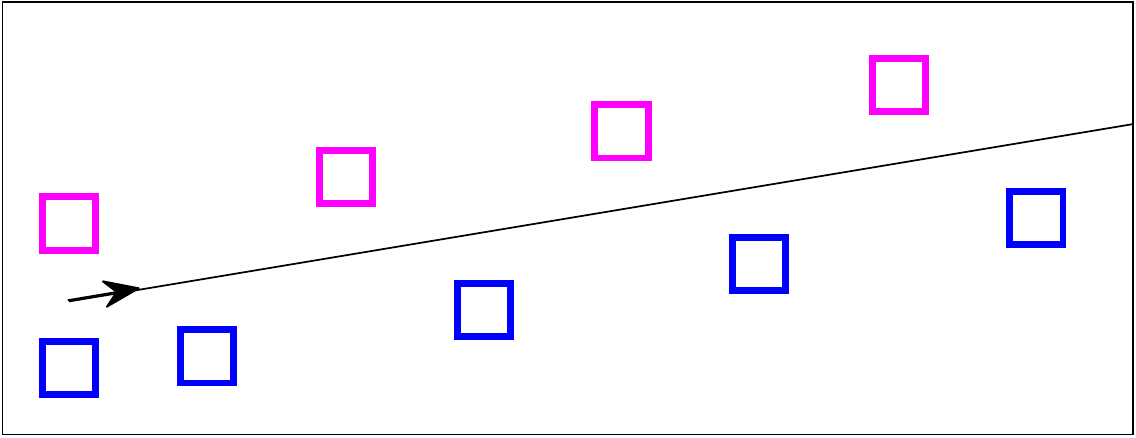}}

\def\irregularPeriodic{\centering\includegraphics[width=\columnwidth]{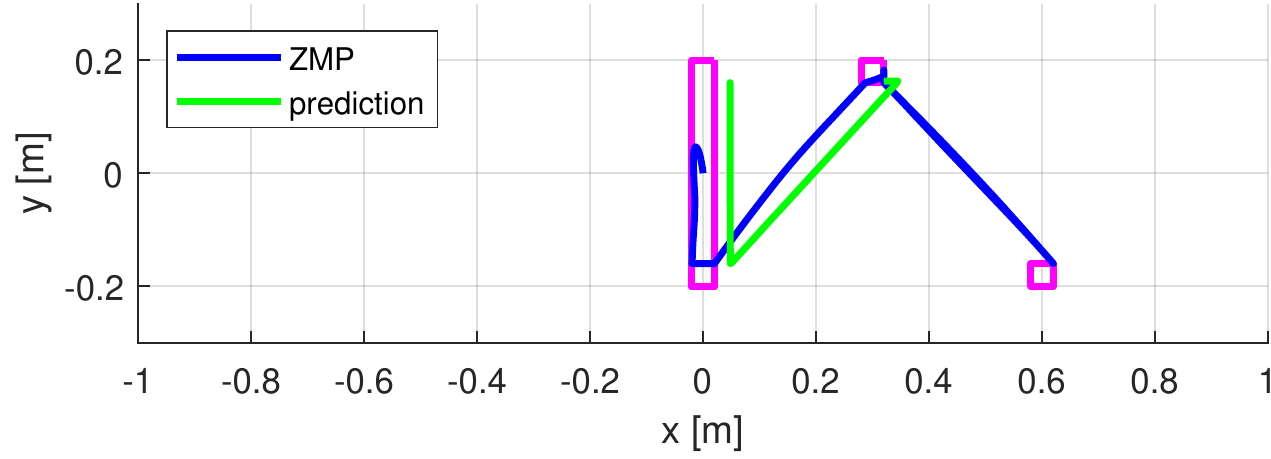}}
\def\irregularPredicted{\centering\includegraphics[width=\columnwidth]{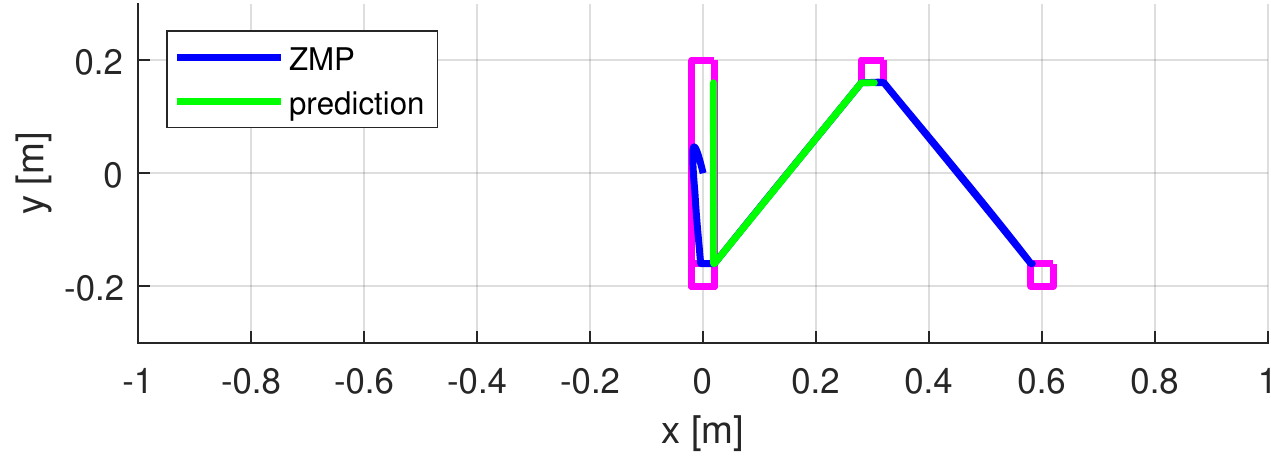}}
\def\regularPeriodic{\centering\includegraphics[width=\columnwidth]{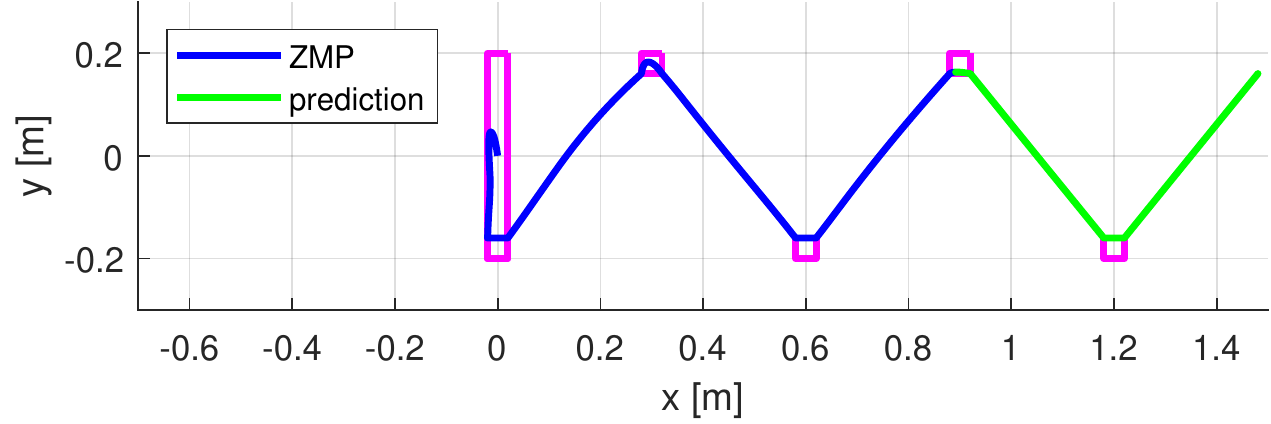}}
\def\regularTruncated{\centering\includegraphics[width=\columnwidth]{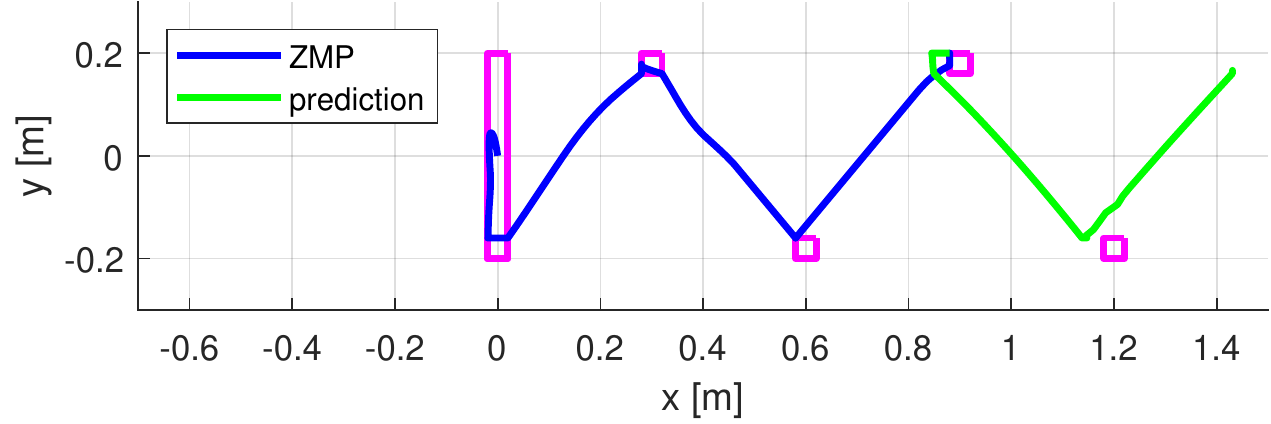}}

\def\stroboStraight{\centering\includegraphics[width=0.9\columnwidth]{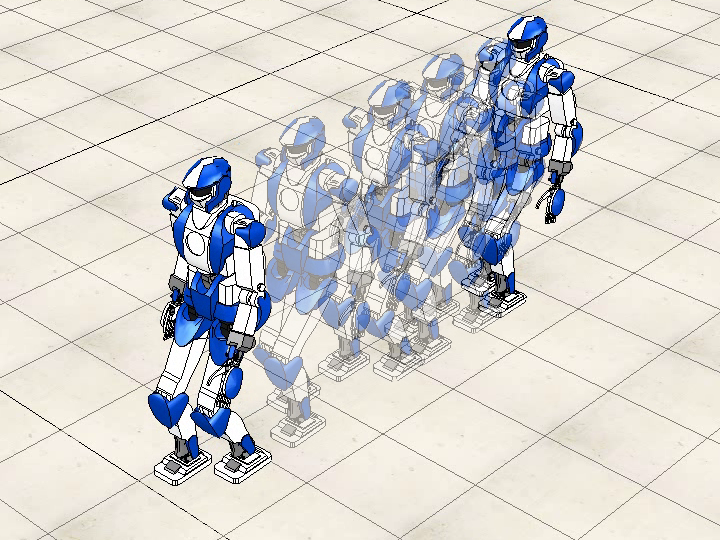}} 
\def\straightWalk{\centering\includegraphics[width=\columnwidth]{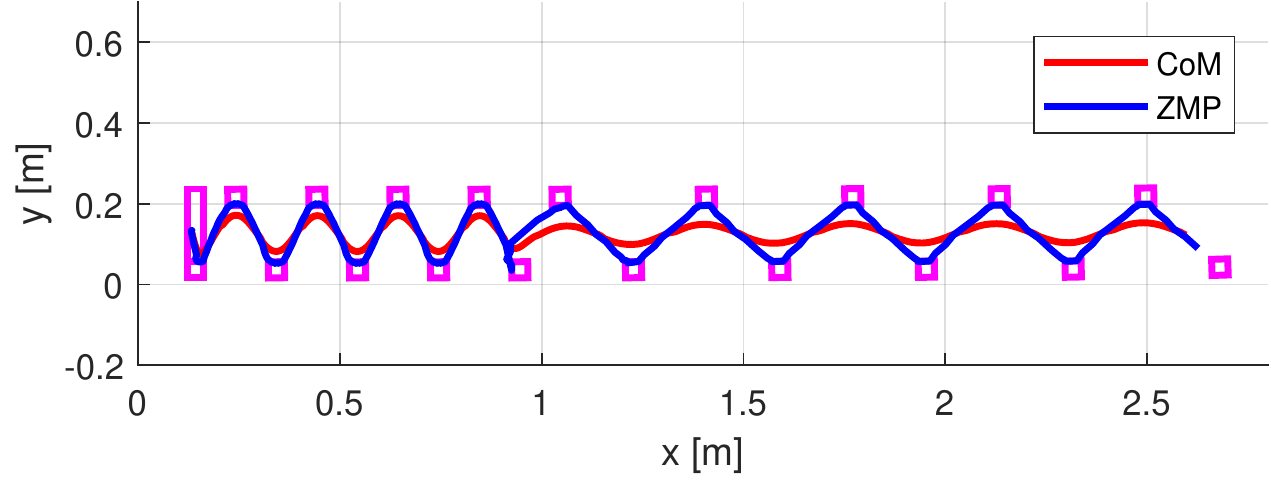}}
\def\refVelocity{\centering\includegraphics[width=\columnwidth]{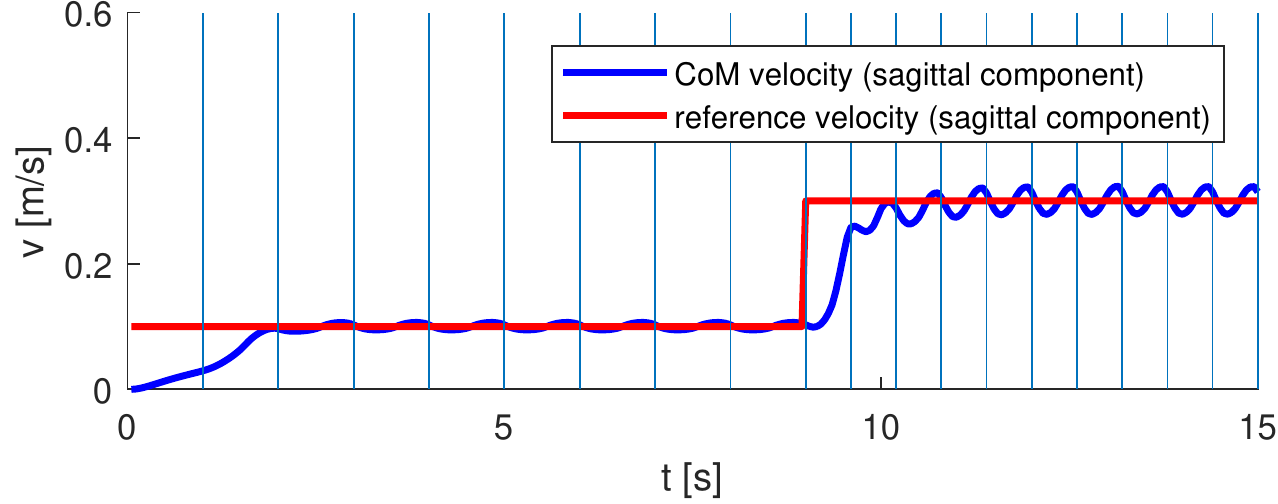}}

\def\stroboCusp{\centering\includegraphics[width=0.9\columnwidth]{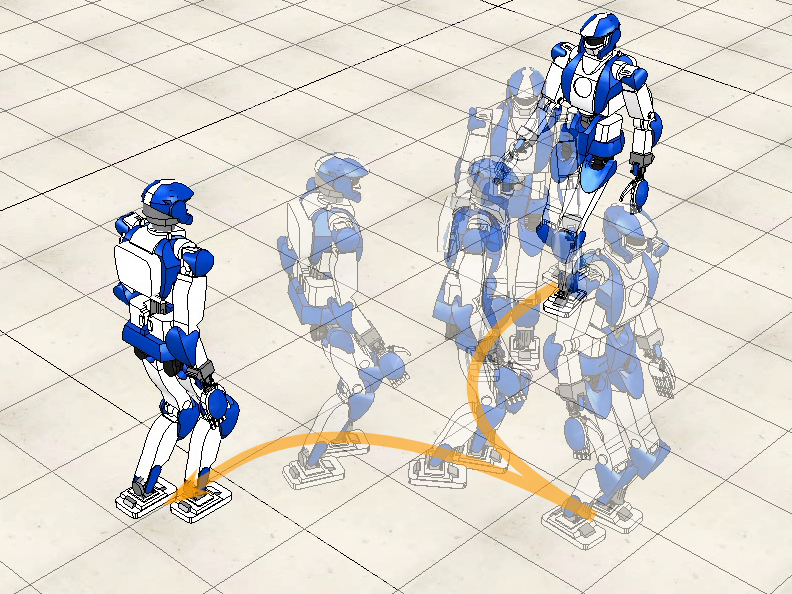}}
\def\CuspWalk{\centering\includegraphics[width=\columnwidth]{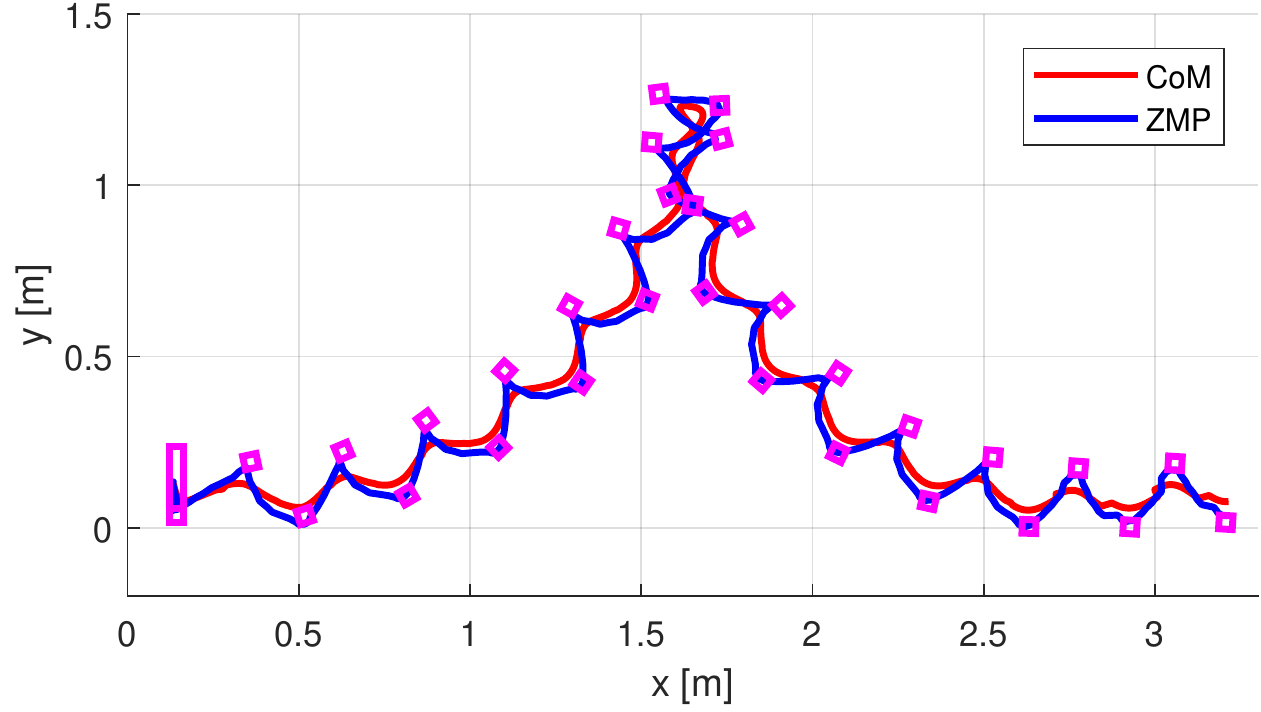}}
\def\refVelocityCusp{\centering\includegraphics[width=\columnwidth]{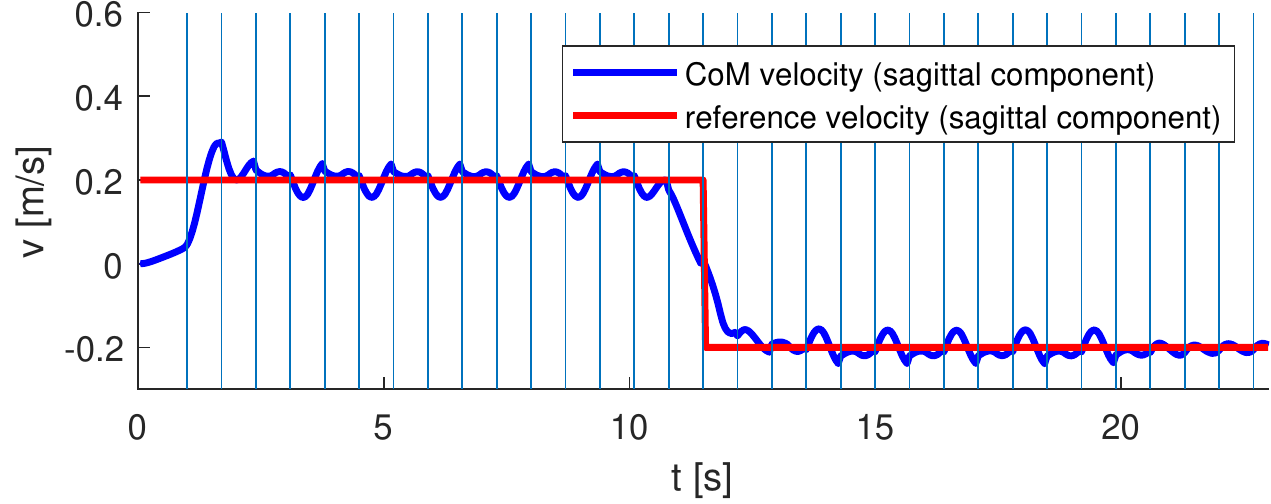}}

\def\Exp_MeasuredZMP{\centering\includegraphics[width=\columnwidth]{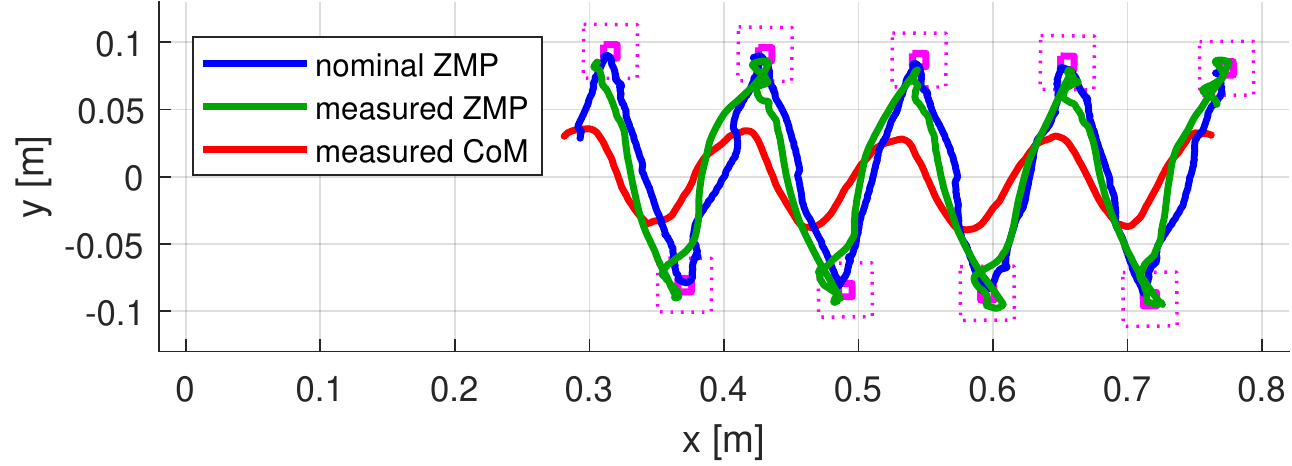}} 

\def\snapshotNAOforwardbackward{\centering\includegraphics[width=\textwidth]{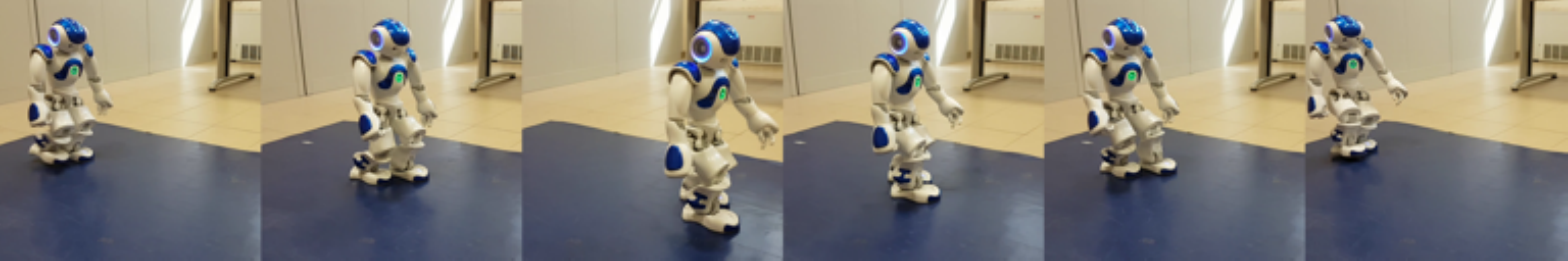}}
\def\snapshotNAOLshape{\centering\includegraphics[width=\textwidth]{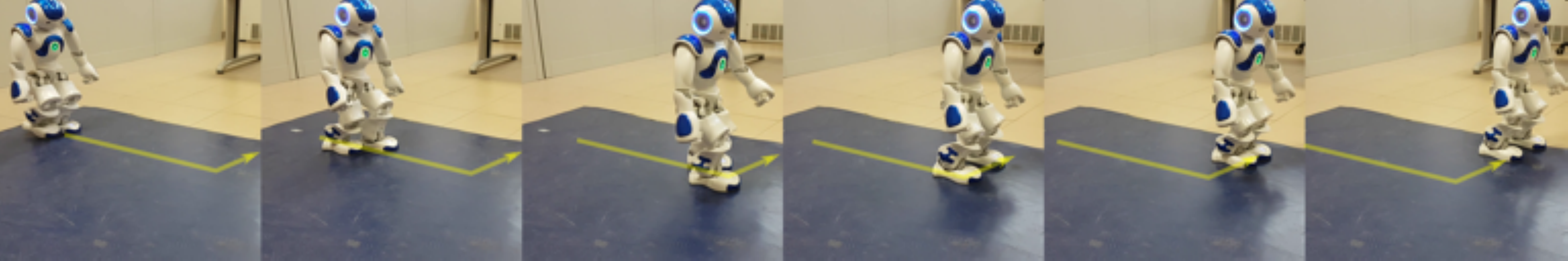}}
\def\snapshotHRPforwardbackward{\centering\includegraphics[width=\textwidth]{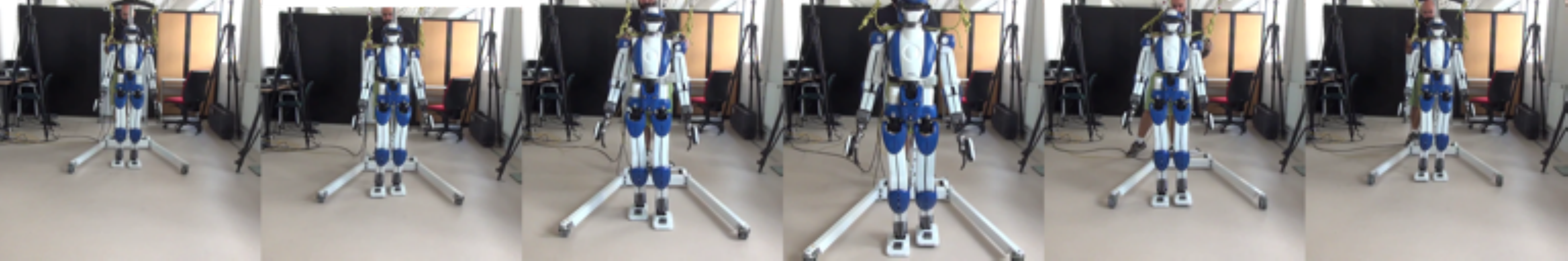}}
\def\snapshotHRPLshape{\centering\includegraphics[width=\textwidth]{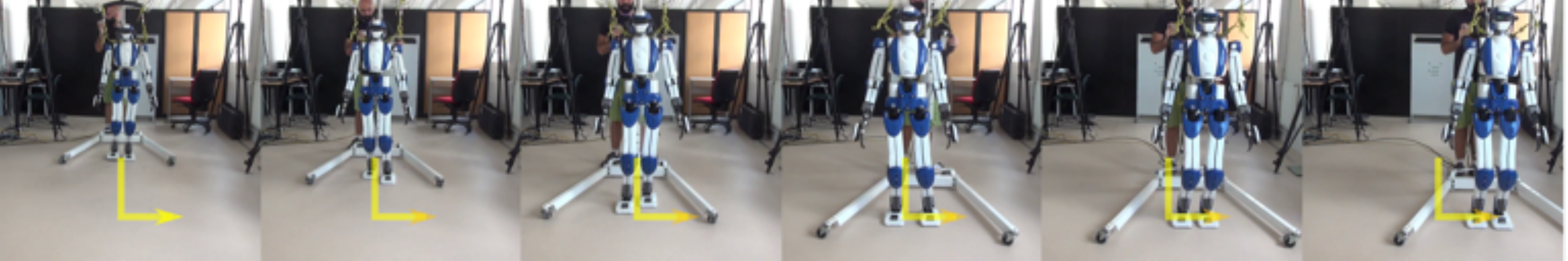}}

\begin{document}

\title{MPC for Humanoid Gait Generation:\\ Stability and Feasibility}

\author{Nicola Scianca, Daniele De Simone, Leonardo Lanari, Giuseppe Oriolo\\
\thanks{The authors are with the Dipartimento di Ingegneria Informatica, Automatica e Gestionale, Sapienza Universit\`a di Roma, Via Ariosto 25, 00185 Rome, Italy. E-mail: \{lastname\}@diag.uniroma1.it. This work was supported by the European Commission through the H2020 project 645097 COMANOID.}}

\maketitle

\begin{abstract} 
We present IS-MPC, an intrinsically stable MPC framework for humanoid gait generation that incorporates a stability constraint in the formulation. The method uses as prediction model a dynamically extended LIP with ZMP velocities as control inputs, producing in real time a gait (including footsteps with timing) that realizes omnidirectional motion commands coming from an external source. The stability constraint links future ZMP velocities to the current state so as to guarantee that the generated CoM trajectory is bounded with respect to the ZMP trajectory. Being the MPC control horizon finite, only part of the future ZMP velocities are decision variables; the remaining part, called {\em tail}, must be either conjectured or anticipated using preview information on the reference motion. Several options for the tail are discussed, each corresponding to a specific terminal constraint. A feasibility analysis of the generic MPC iteration is developed and used to obtain sufficient conditions for recursive feasibility. Finally, we prove that recursive feasibility guarantees stability of the CoM/ZMP dynamics. Simulation and experimental results on NAO and HRP-4 are presented to highlight the performance of IS-MPC.
\end{abstract}



\section{Introduction} 
\label{sect:Introduction}

Many gait generation approaches for humanoids guarantee that balance is maintained during locomotion by enforcing the condition that the Zero Moment Point (ZMP, the point where the horizontal component of the moment of the ground reaction forces becomes zero) remains at all times within the support polygon of the robot. Correspondingly, these approaches identify the ZMP as the fundamental variable to be controlled.

Due to the complexity of full humanoid dynamics, however, direct control of the ZMP is very difficult to achieve. In view of this, simplified models are generally used to relate the evolution of the ZMP to that of the Center of Mass (CoM) of the robot, which can be instead effectively controlled. 
Widely adopted linear models are the Linear Inverted Pendulum (LIP), in which the ZMP represents an input, and the Cart-Table (CT), where the ZMP appears as the output~\cite{KaHiHaYo:14}. The first is appropriate for inversion-based control approaches: given a sequence of footsteps, and thus a ZMP trajectory interpolating them, the LIP is used to compute a CoM trajectory which corresponds to the ZMP trajectory; see, e.g., \cite{HaKaKaHi:06,MoHaKaKaKaFuNaHi:06,BuLoBaUlPf:07}. 
The CT model lends itself more naturally to the design of feedback laws for tracking ZMP trajectories, the most successful example in this context being the LQ preview controller of~\cite{KaKaKaFuHaYoHi:03}.

Regardless of the adopted model, there is a potential instability issue at the heart of the problem. In particular, a certain ZMP trajectory may be realized by an infinity of CoM trajectories, which, due to the nature of the CoM/ZMP dynamics,  will in general be {\em divergent} with respect to the ZMP trajectory itself. In this situation, dynamic balance can be in principle achieved by properly choosing the ZMP trajectory, but {\em internal instability} indicates that such motion will not be feasible in practice for the humanoid.

The seminal paper~\cite{Wi:06} reformulates the gait generation problem in a Model Predictive Control (MPC) setting. This is convenient because it allows to generate simultaneously the ZMP and the CoM trajectories while satisfying constraints, such as the ZMP balance condition as well as kinematic constraints on the maximum step length and foot rotation~\cite{HeDiWiDiMoDi:10}.  Moreover, the MPC approach guarantees a certain robustness against perturbations. It is therefore not surprising that it has been adopted in many methods for gait generation; e.g., see~\cite{AlHeMa:13, FaPoAtIj:14, GrLe:16, FeXiAtKi:16} for linear MPC and~\cite{NaKuStKiMoSo:17,CaKh:17} for nonlinear MPC. 

As for all control schemes, a fundamental issue in MPC approaches is the stability of the obtained closed-loop system, especially in view of the previous remark about the instability of the CoM/ZMP dynamics. As discussed in~\cite{WiTeKu:16}, two main approaches have emerged for achieving stability when MPC is used for humanoid gait generation. The first is heuristic in nature and consists in using a sufficiently long control horizon~\cite{Wi:08}, so that the optimization process can discriminate against diverging behaviors, as done for example in~\cite{HeDiWiDiMoDi:10}. The second approach has been to enforce a terminal state constraint (i.e., a constraint on the state at the end of the control horizon), based on the fact that the MPC literature highlights the beneficial role of such constraints for closed-loop stability in set-point control problems~\cite{MaRaRaSc:00}.  

In particular, terminal constraints were used for humanoid balancing in~\cite{HeOtRo:14} and for gait generation in~\cite{ShDiWi:14}. The latter makes use of a LIP model, requiring its unstable component to stop at the end of the control horizon, a kind of terminal constraint referred to as {\em capturability constraint} (from the concept of capture point~\cite{KoDeReGoPr:12}). 
This constraint has also been used in~\cite{SuYa:17}, where it is imposed only at the foot landing instant, and in~\cite{CaBuMa:17}, which addresses locomotion in a multi-contact setting.

Another approach focusing on the instability issue relies on the concept of Divergent Component of Motion (DCM), used in \cite{TaMaYo:09}  to identify an initial condition for stable execution of regular gaits, and in~\cite{KaKaKuTaShiTaYo:17} to realize transitions between bipedal and quadrupedal gaits. The DCM concept has also been extended to the 3D context in~\cite{EnOtAl:15,CaEsLaMa:20}. More relevant to our review is~\cite{KrEnWiOt:12}, which presents an MPC scheme for gait generation that enforces a terminal constraint (actually converted to a terminal cost for the sake of feasibility) on the DCM component.

In this paper, we move from the fundamental observation that the control problem addressed in MPC-based gait generation is neither a set-point nor a tracking problem. In fact, since the ZMP control objective is encoded via time-varying state constraints, there is no error to be regulated to (or close to) zero. The only significant stability issue in this context is {\em internal stability}, i.e., the boundedness of the CoM trajectory with respect to the ZMP trajectory. Therefore, one cannot simply claim that the use of a terminal constraint will automatically entail internal stability. In fact, to the best of our knowledge, no MPC-based gait generation method exists in the literature for which a rigorous analysis of the stability issue has been performed in connection with the use and the choice of a terminal constraint.

Another tightly related aspect to be considered is that terminal constraints may have a detrimental effect on {\em feasibility}, i.e., the existence of solutions for the optimization problem which is at the core of any MPC scheme~\cite{Ke:01}. A particularly desirable property is {\em recursive feasibility}, which entails that if the optimization problem is feasible at a certain iteration it will remain such in future iterations. It appears that this also crucial issue has seldom been explored for MPC-based gait generation, with the notable exceptions of~\cite{CiWiFr:16,Sh:16}.

In~\cite{ScCoDeLaOr:16} we have introduced a novel MPC approach for humanoid gait generation which relies on the inclusion of an explicit stability constraint in the formulation of the problem. In particular, the idea was to enforce a condition on the future ZMP velocities (representing the control inputs) so as to guarantee that the generated CoM trajectory remains bounded with respect to the ZMP trajectory. Since the control horizon of the MPC algorithm is finite, only part of the future ZMP velocities are decision variables and can therefore be subject to a constraint; the remaining part, called {\em tail}, must be conjectured.

Here, we fully develop our approach into a complete, {\em Intrinsically Stable} MPC (IS-MPC) framework for gait generation. In particular, the paper adds the following contributions with respect to~\cite{ScCoDeLaOr:16}:

\begin{enumerate}

\item we describe a footstep generation module that can be used in conjunction with our MPC scheme in order to modify step timing and length in real time in response to omnidirectional motion commands coming from a higher-level module;

\item depending on the available preview information on the commanded motion, we discuss several versions of the tail (truncated, periodic, anticipative) to be used in the stability constraint, and show that each of them corresponds to a specific terminal constraint; 

\item  we analyze in detail the impact of the new constraint on feasibility, and show analytically how, under certain assumptions, it is possibile to guarantee recursive feasibility of the IS-MPC scheme;  
  
\item we prove that recursive feasibility of IS-MPC implies the desired internal stability of the CoM/ZMP dynamics;

\item  we validate our findings by providing dynamic simulations and actual experiments on two different humanoid robots: an HRP-4 and a NAO.

\end{enumerate}
   
The results on tails, recursive feasibility and internal stability are the main contributions of this paper. We consider them particularly important because they indicate that, contrarily to what is often claimed in the literature, simply adding a terminal constraint (e.g., the capturability constraint) does not {\em per se} guarantee stability of MPC-based gait generation schemes. Indeed, the appropriate tail to be used in the stability constraint --- equivalently, the appropriate terminal constraint --- depends upon the future characteristics of the commanded motion. In this sense, to guarantee recursive feasibility one should always choose the anticipative tail, which makes the most use of the available preview information on such motion. Once recursive feasibility is achieved, CoM/ZMP stability is automatically ensured in IS-MPC.

Another potential benefit of the theoretical analysis of feasibility is that it paves the road for a formal study of the robustness of IS-MPC. Although this is out of the scope of this paper, by relying on this analysis it should be possible to devise modifications of the basic scheme which will preserve recursive feasibility in the presence of quantified bounded uncertainties and/or disturbances.

\begin{figure*}[t]
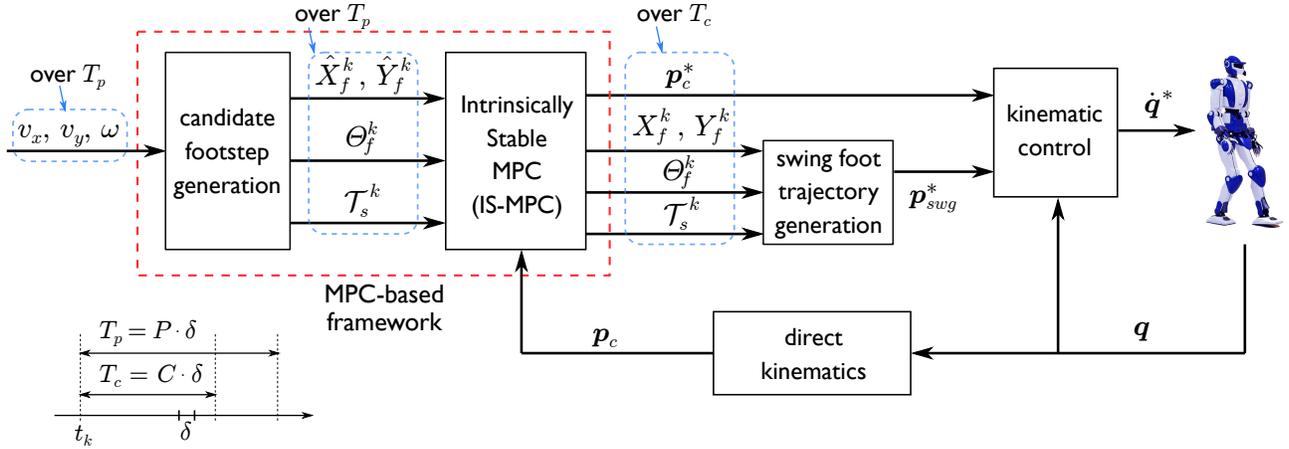

\BlockSchemeDoubleStage
\caption{A block scheme of the proposed MPC-based framework for gait generation.}
\label{fig:BlockSchemeDoubleStage}
\end{figure*}
   
The paper is organized as follows. In the next section, we formulate the considered gait generation problem and discuss the structure of the proposed approach. Section~\ref{sect:FootGen} describes the algorithm which generates timing and locations of the candidate footsteps. In Sect.~\ref{sect:MotModCon} we introduce the prediction model and the constraints used in the IS-MPC scheme, with the exception of the stability constraint which is given a thorough discussion in the dedicated Sect.~\ref{sect:EnforcingStability}. The IS-MPC algorithm is described in detail in Sect.~\ref{sect:ProposedMPC}. Section~\ref{sect:feasibility} addresses the central issues of stability and feasibility of the proposed method; in particular, a theoretical analysis of the feasibility of the generic IS-MPC iteration is presented and used to obtain sufficient conditions for recursive feasibility, whose role in guaranteeing stability is rigorously established. Simulations on the HRP-4 humanoid are presented in Sect.~\ref{sect:Sims}, while experimental results on both the NAO and the HRP-4 humanoids are shown in Sect.~\ref{sect:Exps}. Section~\ref{sect:Conclusions} offers a few concluding remarks.

\section{Problem and Approach} 
\label{sect:ProApp}

Consider the problem of generating a walking gait for a humanoid in response to high-level reference velocities, which are given as the driving ($v_x$, $v_y$) and steering ($\omega$) velocities of an omnidirectional single-body mobile robot chosen as a template model for motion generation.
These velocities, which may encode a persistent trajectory or converge to a stationary point, are produced by an external source; this could be a human operator in a shared control context, or another module of the control architecture working in open-loop (planning) or in closed-loop (feedback control).  

The proposed MPC-based framework, whose block scheme is shown in Fig.~\ref{fig:BlockSchemeDoubleStage}, works in a digital fashion over sampling intervals of duration $\delta$. Throughout the paper, it is assumed that the reference velocities $v_x$, $v_y$, $\omega$ are made available for gait generation with a {\em preview horizon} $T_p=P\cdot \delta$, with $P$ the number of intervals within the preview horizon. At the generic instant $t_k=k\cdot \delta$, the high-level references velocities over $[t_k,t_k+T_p]$ are then sent to the footstep generation module, which uses Quadratic Programming (QP) to generate candidate footsteps over the same interval. In particular, vectors $\hat X_f^k$, $\hat Y_f^k$ collect the Cartesian positions of the footsteps, with the `hat' indicating that these are candidates which can be modified by the MPC module; whereas vector $\Theta_f^k$ collects the footstep orientations, which will not be modified. The footstep generation module also generates the timing ${\cal T}_s^k$ of the sequence.

The output of the footstep generation module is sent to the Intrinsically Stable MPC (IS-MPC) module, which solves another QP problem to produce in real time the actual footstep positions $X_f^k$, $Y_f^k$ and the trajectory $\bfp^\ast_c$ of the humanoid CoM over the {\em control horizon} $T_c=C \cdot \delta$, with $C$ the number of intervals within the control horizon. It is assumed that $T_c \leq T_p$, i.e., $C \leq P$. The inclusion of a stability constraint in the formulation guarantees that the CoM trajectory will be bounded, in a sense to be made precise later. 

The pose (position and orientation) of the footsteps with the associated timing is used to generate --- still in real time --- the swing foot trajectory $\bfp^\ast_{\it swg}$ over the control horizon. Together with the CoM trajectory, this is sent to the kinematic control block, which generates velocity inputs at the joint level in order to achieve output tracking (we are assuming that the humanoid robot is velocity- or position-controlled).

In the next sections we will discuss in detail the proposed control scheme. We will first describe the footstep generation scheme, and then turn our attention to the IS-MPC algorithm, which is our core contribution. The kinematic control block can use any standard pseudoinverse-based feedback law and therefore will not be discussed further.

\section{Candidate Footstep Generation} 
\label{sect:FootGen}

The proposed footstep generation module runs synchronously with the IS-MPC scheme and chooses both the timing and the candidate location of the next  footsteps in response to the high-level reference velocities. Timing is determined first by a simple rule expressing the fact that a change in the reference velocity should affect both the step duration and length. The candidate footstep locations are then chosen through quadratic optimization. 

Note that generating the timing and the orientation of the candidate footsteps outside the IS-MPC is essential to retain the linear structure of the latter. The  IS-MPC scheme will still be able to adapt the position of the footsteps to guarantee reactivity to disturbances. 

At each sampling instant $t_k$, the candidate footstep generation module receives in input the high-level reference velocities over the preview horizon, i.e., from $t_k$ to $t_k+T_p=t_{k+P}$ (see Fig.~\ref{fig:BlockSchemeDoubleStage}). In output, it provides the candidate footstep sequence  $(\hat X^k_f,\hat Y^k_f, \Theta^k_f)$ over the same interval with the associated timing ${\cal T}_s^k$. In particular, these quantities are defined\footnote{To keep a light notation, the $k$ symbol identifying the current sampling instant is used for the sequence vectors but not for their individual elements.} as 
\begin{eqnarray*}
\hat X_f^k &=& (x_f^{1} \>\> \ldots \>\> x_f^{F})^T\\
\hat Y_f^k &=& (y_f^{1} \>\> \ldots \>\> y_f^{F})^T\\
\Theta_f^k &=& (\theta_f^{1} \>\> \ldots \>\> \theta_f^{F})^T
\end{eqnarray*}
and
\[
{\cal T}_s^k = \{T^1_s, \ldots,T^{F}_s\},
\]
where $(x_f^j,y_f^j,\theta_f^j)$ is the pose of the $j$-th footstep in the preview horizon and $T^j_s$ is the duration of the step between the $(j-1)$-th and the $j$-th footstep, taken from the start of the single support phase to the next. Since the duration of steps is variable, the number $F$ of footsteps falling within the preview horizon $T_p$ may change at each $t_k$. 

Below, we discuss first how timing is determined and then describe the procedure for generating the candidate footsteps.

\subsection{Candidate Footstep Timing} 
\label{sect:Timing}

In our method, the duration $T_s$ of each step is related to the magnitude $v=(v_x^2 + v_y^2)^{1/2}$ of the reference Cartesian velocity at the beginning of that step. 

Assume that a triplet of {\em cruise parameters} $(\bar v, \overline{T}_s,\bar L_s)$ has been chosen, where $\bar v$ is a central value of $v$ and $\overline{T}_s$, $\bar L_s$ are the corresponding values of the step duration and length, respectively, with $\bar v = \bar L_s / \overline{T}_s$. The choice of these parameters will depend on the specific kinematic and dynamic capabilities of the humanoid robot under consideration.

The idea is that a deviation from $\bar v$ should reflect on a change in {\em both} $T_s$ and $L_s$. In formulas:
\[
v = \bar v + \Delta v = \frac{\bar L_s + \Delta L_s}{\overline{T}_s - \Delta T_s},
\]
with $\Delta L_s = \alpha \Delta T_s$. 
One easily obtains
\begin{equation}
T_s = \overline{T}_s \, \frac{\alpha + \bar v}{\alpha + v}.
\label{eq:TimingRule}
\end{equation}

\begin{figure}[t]
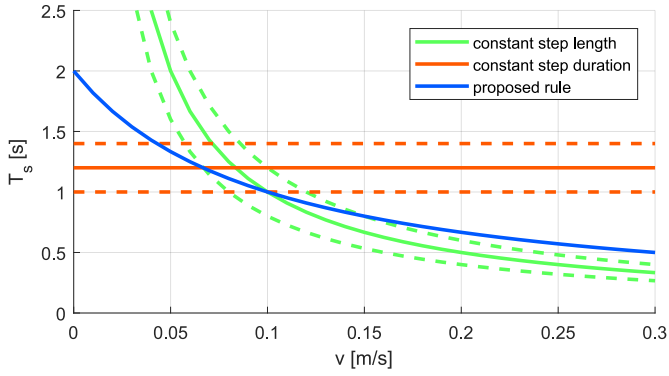

\footstepTimingRule
\caption{The proposed rule for determining the step duration $T_s$ as a function of the magnitude $v$ of the reference Cartesian velocity. For comparison, also shown are the rules yielding constant step duration and constant step length.}
\label{fig:footstepTimingRule}
\end{figure}

Figure~\ref{fig:footstepTimingRule} shows the resulting rule for determining $T_s$ as a function of $v$ in comparison to other possible rules. For illustration, we have set $\bar v= 0.15$~m/s, $\overline{T}_s = 0.8$~s, $\bar L_s=0.12$~m and $\alpha=0.1$~m/s. It is confirmed that an increase of $v$, for example, corresponds to both a decrease of $T_s$ and an increase in $L_s$.

Note that the reference angular velocity $\omega$ does not enter into rule~(\ref{eq:TimingRule}). The rationale is that the step duration and length along curved and rectilinear paths do not differ significantly if the Cartesian velocity $v$ is the same. For a purely rotational motion ($v=0$) where the humanoid is only required to rotate on the spot, the above rule would yield the maximum value of $T_s$.

In practice, equation~(\ref{eq:TimingRule}) is iterated along the preview horizon $[t_k,t_k+T_p]$ in order to obtain the footstep timestamps:
\[
t^j_s = t^{j-1}_s + \overline{T}_s \, \frac{\alpha + \bar v}{\alpha + v(t^{j-1}_s)},
\]
with $t^{0}_s$ equal to the timestamp of the last footstep before $t_k$.
Iterations must be stopped as soon as $t^j_s > t_{k+P}$, discarding the last generated timestamp since it will be outside the preview horizon. The resulting step timing will be ${\cal T}^k_s=\{T^1_s ,\ldots,T^{F}_s\}$, with $T^j_s=t_s^{j+1}-t^{j}_s$.

\subsection{Candidate Footstep Placement}
\label{sect:Footsteps}

Once the timing of the steps in the preview horizon $[t_k,t_k+T_p]$ has been chosen, the poses of candidate footsteps are generated. To this end, we use a reference trajectory obtained by integrating the following template model under the action of the high-level reference velocities over $T_p$:
\begin{equation}
\left( \begin{array}{c}
\dot x\\
\dot y\\
\dot \theta
  \end{array} \right) = 
\left( \begin{array}{ccc}
\cos\theta& -\sin\theta&  0\\
\sin\theta& \cos\theta&  0\\
0& 0 & 1
  \end{array} \right)
\left( \begin{array}{c}
v_x\\
v_y\\
\omega
\end{array} \right).
\label{eq:OmnidirectionalModel}
\end{equation}
This is an omnidirectional motion model which allows the template robot to move along any Cartesian path with any orientation, so as to perform, e.g., lateral walks, diagonal walks, and so on. 

The idea is to distribute the candidate footsteps around the reference trajectory in accordance to the timing ${\cal T}_s^k$ while taking into account the kinematic constraints of the robot. These constraints will also be used in the IS-MPC stage, and therefore we will provide their description directly in Sect.~\ref{sec:KinConst} (see also Fig.~\ref{fig:KinematicConstraint}).

A sequence of two QP problems is solved. The first is
\medskip
\begin{braced}
\[
\min_{\Theta^k_f} \> \sum_{j=1}^{F}(\theta^j_f  -\theta^{j-1}_f  - 
\int_{t^{j-1}_s}^{t^j_s} \omega(\tau)d\tau)^2
\]
\[
\mbox{subject to} \quad  |\theta^j_f-\theta^{j-1}_f| \leq \theta_{\rm max}
\]
\end{braced}
\bigskip

\noindent
Here, $\theta_{\rm max}$ is the maximum allowed rotation between two consecutive footsteps. The second QP problem is

\medskip
\begin{braced}
\[
\min_{\hat X_f^k, \hat Y_f^k} \> \sum_{j=1}^{F} (\hat x_f^j  -\hat x_f^{j-1}  -  \Delta x^j )^2 +
(\hat y_f^j  - \hat y_f^{j-1}  - \Delta y^j )^2 
\]
\centerline{subject to kinematic constraints~(\ref{eq:footposcon})}
\end{braced}
\bigskip

\noindent
Here, $(\hat x_f^0,\hat y_f^0)$ is the known position of the support foot at $t_k$ and $\Delta x^j$, $\Delta y^j$ are given by
\[
\left( \begin{array}{c}
\Delta x^j\\
\Delta y^j
  \end{array} \right) = 
\int_{t^{j-1}_s}^{t^j_s}
R_\theta
\left( \begin{array}{c}
v_x(\tau)\\
v_y(\tau)
\end{array} \right) d\tau \pm
R_j
\left( \begin{array}{c}
0\\
\ell/2
\end{array} \right),
\]
where $R_{\theta}$, $R_j$ are the rotation matrices associated respectively to $\theta(\tau)$ (the orientation of the template robot at any given time $\tau$) and to the footstep orientation $\theta_j$, and $\ell$ is the reference coronal distance between consecutive footsteps. The sign of the second term alternates for left/right footsteps.

At the end of this procedure, the candidate footstep sequence $(\hat X^k_f,\hat Y^k_f, \Theta^k_f)$ with the associated timing ${\cal T}_s^k$ is sent to the IS-MPC stage. The final footstep positions $(X^k_f,Y^k_f)$ will be determined by the latter while the footstep orientations $\Theta^k_f$ and timing ${\cal T}_s^k$ will not be modified.

\begin{figure}[t]
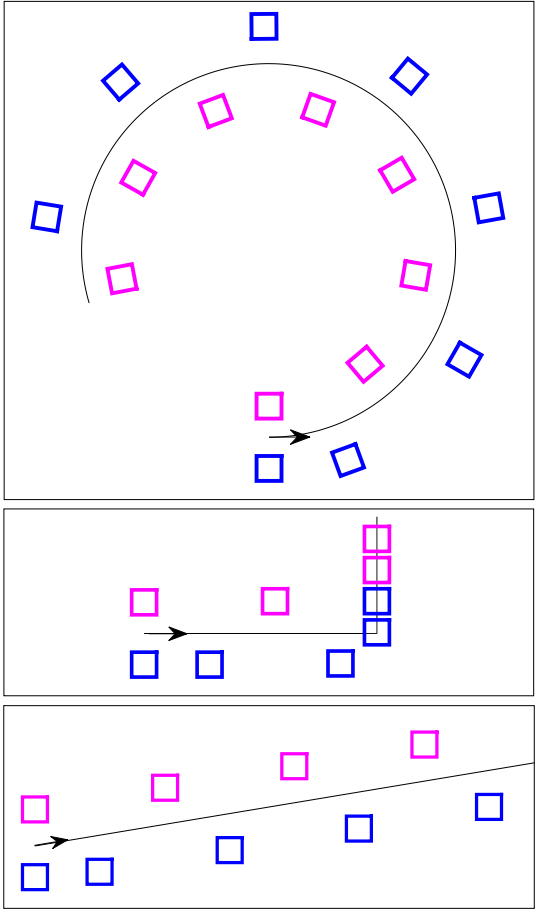

\fsGenCurve \\[0.1cm]
\fsGenLshape \\[0.1cm]
\fsGenDiagonal
\caption{Candidate footsteps generated by the proposed method for different high-level reference velocities corresponding to a circular walk (top), L-walk (center), diagonal walk (bottom). The paths in black are obtained by integrating model~(\ref{eq:OmnidirectionalModel}) under the reference velocities. Footstep in magenta and cyan refer respectively to the left and right foot.}
\label{fig:candidateFootsteps}
\end{figure}

Some examples of candidate footsteps generation are shown in Fig.~\ref{fig:candidateFootsteps}. Note that the orientation of the humanoid robot is tangent to the path for the circular walk, but is kept constant ($\omega=0$) for the other two walks, which represent then proper examples of omnidirectional motion.

\section{IS-MPC: Prediction Model and Constraints} 
\label{sect:MotModCon}

The IS-MPC module uses the Linear Inverted Pendulum (LIP) as a prediction model. The constraints are of three kinds. The first concerns the position of the ZMP, which must be at all times within the support polygon defined by the footstep sequence and the associated timing. The second type of constraint ensures that the generated steps are compatible with the kinematic capabilities of the robot. The third is the new stability constraint guaranteeing that the CoM trajectory generated by our MPC scheme will be bounded with respect to the ZMP trajectory. The first two constraints must be verified throughout the control horizon, whereas the third is a single scalar condition on each coordinate. 

In this section, we discuss in detail the prediction model and the constraints on ZMP and kinematic feasibility. The next section will be devoted to the stability constraint, which deserves a thorough discussion.

\begin{figure}[t]
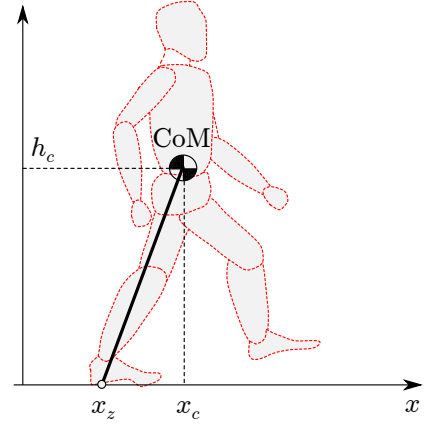

\LIPM_robot
\vspace{-1.2cm}
\caption{The LIP in the $x$ direction.}
\label{fig:LIP}
\end{figure}

\subsection{Prediction Model}
\label{sect:PreMod}

The LIP is a popular choice for describing the motion of the CoM of a biped walking on flat horizontal floor when its height is kept constant and no rotational effects are present. From now on, we express motions in the robot frame, which has its origin at the center of the current support foot, the $x$-axis {\em (sagittal)} aligned with the support foot, and the $y$-axis {\em (coronal)} orthogonal to the $x$-axis. In the LIP model, which applies to both point feet and finite-sized feet, the dynamics along the sagittal and coronal axes are governed by decoupled, identical linear differential equations.

Consider for illustration the motion along the $x$ axis (see Fig.~\ref{fig:LIP}), and let  $x_c$ and $x_z$ be respectively the coordinate of the CoM and the ZMP. The LIP dynamics is 
\begin{equation}
\ddot x_c = \eta^2 (x_c - x_z),
\label{eq:LIP_eq}
\end{equation}
where $\eta = \sqrt{g/h_c}$, with $g$ the gravity acceleration and $h_c$ the constant height of the CoM. In this model, the ZMP position $x_z$ represents the input, whereas the CoM position $x_c$ is the output.

To obtain smoother trajectories, we take the ZMP velocity $\dot x_z$ as the actual control input. 
This leads to the following third-order prediction model (LIP $+$ dynamic extension)
\begin{equation}
\left( \begin{array}{c}
\dot x_c\\
\ddot x_c\\
\dot x_z
  \end{array} \right) = 
\left( \begin{array}{ccc}
0& 1&  0\\
\eta^2& 0&  -\eta^2\\
0& 0 & 0
  \end{array} \right)
\left( \begin{array}{c}
x_c\\
\dot x_c\\
x_z
\end{array} \right) + \left( \begin{array}{c} 0 \\ 0 \\ 1 \end{array}\right) \dot x_z.
\label{eq:ThirdOrder}
\end{equation}

\begin{figure}[t]
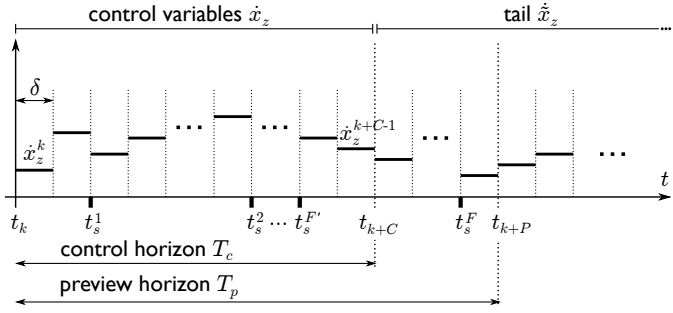

\PiecewiseConstZMPVel
\medskip
\caption{At time $t_k$, the control variables determined by IS-MPC are the piecewise-constant ZMP velocities over the control horizon. The ZMP velocities after the control horizon are instead conjectured in order to build the tail (see Sect.~\ref{sect:Tails}). Also shown are the $F$ footstep timestamps placed by the footstep generation module in the preview horizon; $F'$ of them fall in the control horizon.}
\label{fig:PiecewiseConstZMPVel}
\end{figure}

Our MPC scheme uses piecewise-constant control over the sampling intervals (see Fig.~\ref{fig:PiecewiseConstZMPVel}):
\[
\dot x_z(t) = \dot x_z^i, \quad t \in [t_i, t_{i+1}).
\] 
In particular, a bound of the form $\vert \dot x_z^i \vert \leq \gamma$, with $\gamma$ a positive constant, will be satisfied for all $i$. In fact, the reference velocities $v_x$, $v_y$, $\omega$ will be bounded in any realistic gait generation problem. As shown by Fig.~\ref{fig:footstepTimingRule}, the footstep generation module will then produce a sequence of footsteps along which the step duration is bounded below. This timing will be reflected in the associated ZMP constraints (see Sect.~\ref{sect:ZMPcon}), which will in turn entail as solution a piecewise-continuous trajectory $x_z(t)$ with bounded derivative.
Therefore, for $t \in [t_i, t_{i+1}]$ it will be
\begin{equation}
x_z (t)  =  x_z^ i  + (t - t_i) \, \dot x_z^i, \quad \mbox{with } \vert \dot x_z^i \vert \leq \gamma, 
\label{eq:LinearZ}
\end{equation}
where we have used the notation $x_z^{i}=x_z(t_i)$.

The generic iteration of IS-MPC plans over the control horizon, i.e., from $t_k$ to $t_k+T_c=t_{k+C}$. Since $T_c \leq T_p$, a subset of the $F$ candidate footsteps produced by the footstep generation module fall inside the control horizon; denote their number by $F'<F$. 
The MPC iteration will then generate:

\begin{itemize} 

\item[$\bullet$] the control variables, i.e., the input values $\dot x_z^{k+i}$, $\dot y_z^{k+i}$, for $i=0,\dots,C-1$; 

\item[$\bullet$] the other decision variables, i.e., the actual footstep positions $(x_f^j , x_f^j)$, for $j\!=\!1,\ldots,F'$. 

\item[$\bullet$] as a byproduct, the output history $x_c(t)$, $y_c(t)$, for $t \in [t_k,t_{k+C}]$ which will be ultimately used to drive the actual humanoid.

\end{itemize}

As already mentioned, the orientations of the footsteps are instead inherited from the generated sequence (more on this in Sect.~\ref{sect:ZMPcon}).

Note that the footsteps do not appear in the prediction model, but will show up in the constraints, as discussed in the rest of this section.

\subsection{ZMP Constraints}
\label{sect:ZMPcon}

The first constraint guarantees dynamic balance by imposing that the ZMP lies inside the current support polygon at all time instants within the control horizon. 

When the robot is in single support on the $j$-th footstep, the admissible region for the ZMP is the interior of the footstep, which can be approximated as a rectangle of dimensions $d_{z,x}, d_{z,y}$, centered at $(x_f^j , y_f^j)$, and oriented as $\theta^j$. Using the fact that the ZMP profile is piecewise-linear as entailed by~(\ref{eq:LinearZ}), the constraint can be expressed as\footnote{For compactness, we shall only write the right-hand side of bilateral inequality constraints. For example, constraint~(\ref{eq:ZMPcon}) should be completed by a left-hand side obtained by adding (rather than subtracting) the two terms that appear in the right-hand side.}:
\begin{equation}
R_j^T \left( \begin{array}{c} \delta\sum_{l=0}^{i}\dot x_z^{k+l} - x_f^j \\ [10pt]
\delta\sum_{l=0}^{i}\dot y_z^{k+l} - y_f^j \end{array}  \right) \leq
\frac{1}{2}\left(  \begin{array}{c} d_{z,x} \\  [5pt]
d_{z,y} \end{array}  \right) -
R_j^T \left(  \begin{array}{c} x_z^{k} \\ [5pt]
y_z^{k} \end{array}  \right).
\label{eq:ZMPcon}
\end{equation}
If the above sampled-time ZMP constraint is satisfied, then the original continuous-time constraint is also satisfied thanks to the linearity of $x_z(t)$ within each sampling interval. Constraint~(\ref{eq:ZMPcon}), complete with the corresponding left-hand side, must be imposed throughout the control horizon ($i=0,\dots,C-1$) and for all the associated footsteps ($j=0,\dots,F'$). 

Note that constraint~(\ref{eq:ZMPcon}) is nonlinear in the footstep orientation $\theta^j$, which however is not a decision variable, being simply inherited from the footstep generation module. The constraint is instead linear in $x_f^j$, $y_f^j$, as well as in the ZMP velocity inputs.

\begin{figure}[t]
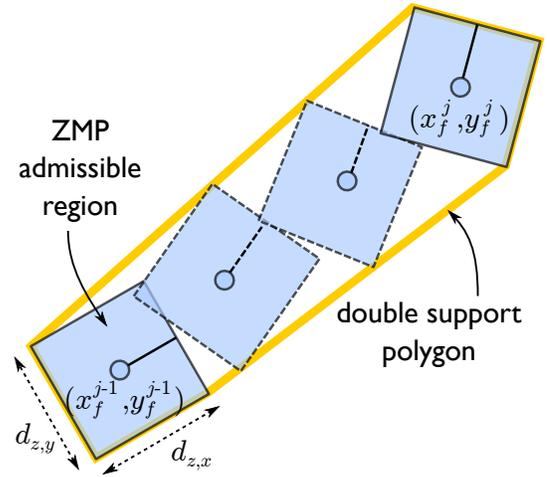

\DoubleSupport
\medskip
\caption{The ZMP moving constraint in double support.}
\label{fig:DoubleSupport}
\end{figure}

During double support, the support polygon would be the convex hull of the two footsteps, whose boundary is a nonlinear function of their relative position. To preserve linearity, we adopt an approach based on {\em moving constraints} \cite{AbScDeLaOr:17}. In particular, the admissible region for the ZMP in double support has exactly the same shape and dimensions it has in single support, and it roto-translates (i.e., simultaneously rotates and translates) from one footstep to the other in such a way to always remain in the support polygon (see Fig.~\ref{fig:DoubleSupport}). This results in a slightly conservative constraint which is however linear in the decision variables.

\subsection{Kinematic Constraints}
\label{sec:KinConst}

\begin{figure}[t]
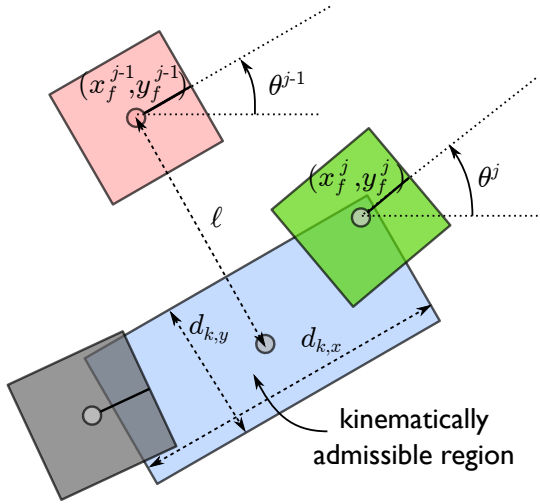

\KinematicConstraint
\medskip
\caption{The kinematic constraint on footstep placement.}
\label{fig:KinematicConstraint}
\end{figure}

The second type of constraint is introduced to ensure that all steps are compatible with the robot kinematic limits. Consider the $j$-th step in $T_c$, with the support foot centered at $(x_f^{j-1} , y_f^{j-1})$ and oriented as $\theta^{j-1}$. The admissible region for placing the footstep is defined as a rectangle having the same orientation $\theta^{j-1}$ and whose center is displaced from the support foot center by a distance $\ell$ in the coronal direction (see Fig.~\ref{fig:KinematicConstraint}). Denoting by $d_{a,x}$ and $d_{a,y}$ the dimensions of the kinematically admissible region, the constraint can be written as
\begin{equation}
R_{j-1}^T \left(  \begin{array}{c} x_f^{j} - x_f^{j-1} \\ [5pt]
 y_f^{j} - y_f^{j-1} \end{array}  \right) \leq
 \pm \left(  \begin{array}{c} 0 \\ [5pt]
\ell \end{array}  \right) + \frac{1}{2}
\left(  \begin{array}{c} d_{a,x} \\ [5pt]
d_{a,y} \end{array}  \right),
\label{eq:footposcon}
\end{equation}
with the sign alternating for the two feet. The above constraint, complete with the corresponding left-hand side, must be imposed for all footsteps in the control horizon ($j=1,\dots,F'$).

\section{IS-MPC: Enforcing Stability}
\label{sect:EnforcingStability}

The LIP dynamics (\ref{eq:LIP_eq}) is inherently unstable. As a consequence, even when the ZMP lies at all times within the support polygon {\em (gait balance)} it may still happen that the CoM diverges exponentially with respect to the ZMP; in this case, the gait would obviously become unfeasible in practice, due to the kinematic limitations of the robot. The role of the stability constraint is then to guarantee that the CoM trajectory remains bounded with respect to the ZMP {\em (internal stability)}.

In this section, we first describe the structure of the stability constraint and then discuss the possible {\em tails} for its implementation.

\subsection{Stability Constraint}
\label{sect:StabilityConstraint}

Since we want to enforce boundedness of the CoM w.r.t.\ the ZMP, we can ignore the dynamic extension and focus directly on the LIP system.
 
By using the following change of coordinates
\begin{eqnarray}
x_s &=& x_c - \dot x_c/\eta \label{eq:xstransf}\\
x_u &=& x_c + \dot x_c/\eta, \label{eq:xutransf}
\end{eqnarray}
the LIP part of system~(\ref{eq:LIP_eq}) is decomposed into a stable and an unstable subsystem:
\begin{eqnarray}
\dot x_s &=& - \eta\,(x_s - x_z) \label{eq:xsdyn}\\
\dot x_u &=& \eta\,(x_u - x_z). \label{eq:xudyn}
\end{eqnarray}
 The unstable component $x_u$ is also known as {\em divergent component of motion} (DCM) \cite{TaMaYo:09} or {\em capture point} \cite{PrCaDrGo:06}.
 
In spite of the LIP instability, for any input ZMP trajectory $x_z(t)$ of the form~(\ref{eq:LinearZ}) there exists a special initialization of $x_u$ such that the resulting output CoM trajectory is bounded with respect to the input~\cite{LaHu:15}. In particular, this is the (only) initial condition on $x_u$ for which the free evolution of~(\ref{eq:xudyn}) exactly cancels the component of the forced evolution that would diverge with respect to $x_z(t)$. In the MPC context, where the initial condition at $t_k$ is denoted by $x_u(t_k)=x_u^k$, the special initialization is expressed as
\begin{equation}
x_u^k =  \eta \int_{t_k}^\infty e^{-\eta(\tau-t_k)} x_z(\tau) d\tau.
\label{eq:bcConstr}
\end{equation}
Note that this particular initialization depends on the future values of the LIP input, i.e., the ZMP coordinate $x_z$. In the following, we refer to~(\ref{eq:bcConstr}) as the {\em stability condition}. 

The stability condition, which involves $x_u$ at the initial instant $t_k$ of the control horizon, can be propagated to its final instant $t_{k+C}$ by integrating~(\ref{eq:xudyn}) from $x_u^k$ in~(\ref{eq:bcConstr}):
\begin{equation}
x_u^{k+C} = \eta \int_{t_{k+C}}^\infty e^{-\eta(\tau - t_{k+C})}x_z(\tau)d\tau.
\label{eq:TC}
\end{equation}

Condition~(\ref{eq:bcConstr}) --- or equivalently,~(\ref{eq:TC}) --- can be used to set up the corresponding constraint for the MPC problem. To this end, we use the piecewise-linear profile~(\ref{eq:LinearZ}) of $x_z$ to obtain explicit forms.

\medskip

\begin{proposition}
\label{prop:first}
For the piecewise-linear $x_z$ in (\ref{eq:LinearZ}), condition~(\ref{eq:bcConstr}) becomes
\begin{equation}
x_u^k = x_z^k + \frac{1-e^{-\eta\delta}}{\eta}\sum\limits_{i=0}^\infty e^{-i\eta\delta}\dot x_z^{k+i},
\label{eq:GenConstraint}
\end{equation}
while~(\ref{eq:TC}) takes the form
\begin{equation}
x_u^{k+C} = x_z^{k+C} + \frac{1-e^{-\eta\delta}}{\eta} e^{C\eta\delta}\sum_{i=C}^\infty e^{-i\eta\delta}\dot x_z^{k+i}.
\label{eq:GenTerConstraint}
\end{equation}
\end{proposition}

\medskip

{\em Proof}.
Rewrite eq.~(\ref{eq:LinearZ})  as
\begin{equation}
x_z(t) = x_z^k + \sum\limits_{i=0}^\infty (\rho (t-t_{k+i}) - \rho (t-t_{k+i+1})) \dot x_z^{k+i},
\label{eq:xzramp}
\end{equation}
where $\rho(t) = t \,\delta_{-1}(t)$ denotes the unit ramp and $\delta_{-1}(t)$  the unit step. Using Properties~1, 4 and 3 given in the Appendix, we get
\[
\int_{t_k}^\infty \!\!\!\!\! e^{-\eta (\tau -t_k)}(\rho (\tau-t_{k+i}) - \rho (\tau-t_{k+i+1})) d \tau \!=\! \frac{1\!-\!e^{-\eta \delta}}{\eta^2}e^{-i\eta \delta}.
\]
Plugging this expression in condition~(\ref{eq:bcConstr}) and using Property 2 of the Appendix one obtains~(\ref{eq:GenConstraint}).

 To prove~(\ref{eq:GenTerConstraint}), rewrite~(\ref{eq:xzramp}) as
\begin{align}
x_z(t) =  x_z^k +& \sum_{i=0}^{C-1} (\rho (t-t_{k+i}) - \rho (t-t_{k+i+1})) \dot x_z^{k+i} \nonumber \\
+& \sum_{i=C}^\infty (\rho (t-t_{k+i}) - \rho (t-t_{k+i+1})) \dot x_z^{k+i}. \nonumber
\end{align}
The contribution of the first two terms of $x_z$ to the integral in~(\ref{eq:TC}) is $x_z^{k+C}$.  Using Properties~1, 3 and 4 one verifies that the contribution of the third term is exactly the second term in the right hand side of~(\ref{eq:GenTerConstraint}). This completes the proof.\hfill\bull

\medskip

In~(\ref{eq:GenConstraint}), one should logically separate the values of $\dot x_z^{i}$ within the control horizon, i.e. the control variables $\dot x_z^{i}$ for $i = k, \dots, k+C-1$, from the remaining values, i.e., from $k+C$ on. 
The infinite summation is then split in two parts and (\ref{eq:GenConstraint}) can be rearranged as\footnote{Constraint~(\ref{eq:StabConstrSplit}) can be written as a function of the actual state variables of our prediction model ($x_c$, $\dot x_c$ and $x_z$) using the coordinate transformation~(\ref{eq:xutransf}). The same is true for all subsequent forms of the stability constraint as well as of the terminal constraint.}
\begin{equation}
\sum_{i=0}^{C-1} e^{-i\eta\delta}\dot x_z^{k+i} = -\sum_{i=C}^\infty e^{-i\eta\delta}\dot x_z^{k+i} + 
\frac{\eta}{1-e^{-\eta\delta}}(x_u^k - x_z^k).
\label{eq:StabConstrSplit}
\end{equation}
Observe the inversion between (\ref{eq:GenConstraint}), which expresses the stable initialization at $t_k$ for a given $x_z(t)$, and (\ref{eq:StabConstrSplit}), which constrains the control variables so that the associated stable initialization matches the current state at $t_k$. In the following, we will refer to~(\ref{eq:StabConstrSplit}) as the {\em stability constraint}.

The control variables do not appear  in condition~(\ref{eq:GenTerConstraint}), which involves only the value of the state variable $x_u^{k+C}$ at the end of the control horizon. In other terms, this condition represents what is called a {\em terminal constraint} in the MPC literature.

Both the stability and the terminal constraint contain an infinite summation which depends on $\dot x_z^{k+C}$, $\dot x_z^{k+C+1}, \dots$, i.e., the ZMP velocities {\em after} the control horizon. These are obviously unknown, because they will be determined by future iterations of the MPC algorithm; as a consequence, including either of the constraints in the MPC formulation would lead to a non-causal (unrealizable) controller. 
However, by exploiting the preview information on $v_x$, $v_y$, $\omega$, we can make an {\em informed conjecture} at $t_k$ about these ZMP velocities, which we will denote by $\dot{\tilde x}_z^{k+C}$, $\dot{\tilde x}_z^{k+C+1}, \dots$ and refer to collectively as the {\em tail} in the following. Correspondingly, the stability constraint~(\ref{eq:StabConstrSplit}) assumes the form
\begin{equation}
\sum_{i=0}^{C-1} e^{-i\eta\delta}\dot x_z^{k+i} = -\sum_{i=C}^\infty e^{-i\eta\delta} \dot{\tilde x}_z^{k+i} + 
\frac{\eta}{1-e^{-\eta\delta}}(x_u^k - x_z^k),
\label{eq:StabConstrSplitCaus}
\end{equation}
while the terminal constraint~(\ref{eq:GenTerConstraint}) becomes
\begin{equation}
x_u^{k+C} = x_z^{k+C} + \frac{1-e^{-\eta\delta}}{\eta} e^{C\eta\delta}\sum_{i=C}^\infty e^{-i\eta\delta} \dot{\tilde x}_z^{k+i}.
\label{eq:GenTerConstraintCaus}
\end{equation}
Using either of these in the MPC formulation will lead to a causal (realizable) controller. 
 
\subsection{Tails}
\label{sect:Tails}

We now discuss three possible options for the structure of the tail depending on the assumed behavior of the ZMP velocities after the control horizon. Basically, they correspond to {\em (i)} neglecting them {\em (ii)} assuming they are periodic {\em (iii)} anticipating a more general profile based on preview information. For each option, we shall explicitly compute the corresponding form of both the stability and the terminal constraint.

\medskip

\subsubsection{Truncated Tail}

The simplest option is to {\em truncate} the tail, by assuming that the corresponding ZMP velocities are all zero. This is a sensible choice if the preview information indicates that the robot is expected to stop at the end of the control horizon.

\medskip

\begin{proposition}
Let {\em (truncated tail)}
\[
\dot {\tilde x}_z^{k+i}=0  \qquad \mbox{for} \quad i \geq C.
\]
The stability constraint becomes
\begin{equation}
\sum_{i=0}^{C-1} e^{-i\eta\delta}\dot x_z^{k+i} = 
\frac{\eta}{1-e^{-\eta\delta}}(x_u^k - x_z^k),
\label{eq:SC_Trunc}
\end{equation}
while the terminal constraint becomes
\begin{equation}
x_u^{k+C} = x_z^{k+C}.
\label{eq:TC_Trunc}
\end{equation}
\end{proposition}

\medskip

{\em Proof}.
The above expressions are readily derived from the general constraints~(\ref{eq:StabConstrSplitCaus}) and~(\ref{eq:GenTerConstraintCaus}), respectively.\hfill\bull

\medskip

Interestingly, the terminal constraint~(\ref{eq:TC_Trunc}) is equivalent to the {\em capturability constraint}, originally introduced in~\cite{ShDiWi:14}.

\medskip

\subsubsection{Periodic Tail}
\label{sect:PerTail}

The second option is to use a {\em periodic tail} obtained by infinite replication of the ZMP velocities within the control horizon. This assumption is justified when the reference velocities are themselves periodic (in particular, constant) in $T_c$, which is typically chosen as the gait period (total duration of two consecutive steps) or a multiple of it. Formulas for a replication period different from the control horizon may be easily derived. 

\medskip

\begin{proposition}
Let {\em (periodic tail)}
\begin{eqnarray*}
\dot{\tilde x}_z^{k+i} &=& \dot x_z^{k+i-C} \qquad \mbox{for} \quad i=C,\dots, 2C-1\\
\dot{\tilde x}_z^{k+i} &=& \dot{\tilde x}_z^{k+i-C} \qquad \mbox{for} \quad i \geq 2C. 
\end{eqnarray*}
The stability constraint becomes
\begin{equation}
\sum_{i=0}^{C-1} e^{-i\eta\delta}\dot x_z^{k+i} =  \eta \, \frac{1- e^{-C\eta\delta}}{1-e^{-\eta\delta}}(x_u^k - x_z^k),
\label{eq:SC_Per}  
\end{equation}
while the terminal constraint becomes 
\begin{equation}
x^{k+C}_u - x^{k+C}_z = x^{k}_u - x^{k}_z.
\label{eq:TC_Per}
\end{equation}
\end{proposition}

\medskip

{\em Proof}.
 If the tail is periodic, the infinite summation in~(\ref{eq:StabConstrSplitCaus}) can be rewritten as follows:
\begin{align}
\sum_{i=C}^\infty e^{-i\eta\delta}\dot{\tilde x}_z^{k+i} &= e^{-C\eta\delta}\sum_{i=0}^\infty e^{-i \eta \delta} \dot{\tilde x}_z^{k+C+i} \nonumber\\
&= e^{-C\eta\delta} \sum_{i=0}^{C-1} e^{-i \eta \delta} \dot x_z^{k+i}\left( 1 + e^{-C\eta \delta} +  \dots \right)\nonumber\\
&= \frac{e^{-C\eta\delta}}{1 - e^{-C\eta\delta}}\sum_{i=0}^{C-1} e^{-i \eta \delta} \dot x_z^{k+i},\nonumber
\end{align}
which can be plugged in~(\ref{eq:StabConstrSplitCaus}) and in~(\ref{eq:GenTerConstraintCaus}), respectively, to obtain~(\ref{eq:SC_Per}) and~(\ref{eq:TC_Per}).
\hfill\bull
\medskip

Note that, using~(\ref{eq:xudyn}), the terminal constraint (\ref{eq:TC_Per}) can be rewritten as 
\[
\dot x^{k+C}_u = \dot x^{k}_u.
\]

\medskip

\subsubsection{Anticipative Tail}
\label{sect:PredTail}

In the general case, one can use the candidate footsteps produced by the footstep generation module {\em beyond} the control horizon to conjecture a tail  in $[T_c, T_p]$. This is done in two phases: in the first, we generate in $[T_c, T_p]$ a ZMP trajectory which belongs at all times to the admissible ZMP region defined by the footsteps $\{(\hat x^{F'}_f,\hat y^{F'}_f, \theta^{F'}_f), \dots,  (\hat x^{F}_f,\hat y^{F}_f, \theta^{F}_f)\}$. In the second phase, we sample the time derivative of this ZMP trajectory every $\delta$ seconds.

Denote the samples obtained by the above procedure by $\dot x_{z,\rm ant}^{k+i}$, for $i=C,\dots,P-1$. The {\em anticipative tail} is then obtained by: 

\begin{itemize}

\item setting $\dot{\tilde x}_z^{k+i} = \dot x_{z,\rm ant}^{k+i}$ for $i=C,\dots,P-1$; 

\item using a truncated or periodic expression for the residual part of the tail located {\em after} the preview horizon, i.e., for $\dot {\tilde x}_z^{k+i}$, $i=P,P+1,\dots \,$.

\end{itemize}

The stability constraint (\ref{eq:StabConstrSplitCaus}) then becomes 
\begin{align*}
\sum_{i=0}^{C-1} e^{-i\eta\delta}\dot x_z^{k+i} = -&\sum_{i=C}^{P-1} e^{-i\eta\delta}\dot x_{z,\rm ant}^{k+i} -\sum_{i=P}^{\infty} e^{-i\eta\delta}\dot {\tilde x}_z^{k+i} +\\
&  \frac{\eta}{1-e^{-\eta\delta}}(x_u^k - x_z^k).
\end{align*}
Once a form is chosen for the residual part of the tail, this formula leads to a closed-form expression of the stability constraint which consists of a finite number of terms, and is therefore still amenable to real-time implementation. Similarly, one can use~(\ref{eq:GenTerConstraintCaus}) to derive the corresponding expression of the terminal constraint.

In the following, and specifically in the feasibility analysis of Sect.~\ref{sect:Recursive}, we will use a particular form of anticipative tail such that {\em (i)} the ZMP trajectory in $[T_c, T_p]$ is always at the center of the ZMP admissible region, and {\em (ii)} the residual part of the tail is truncated.

\section{IS-MPC: Algorithm} 
\label{sect:ProposedMPC}

Each iteration of our IS-MPC algorithm solves a QP problem based on the prediction model and constraints described in Sect.~\ref{sect:MotModCon}, with the addition of the stability constraint discussed in the previous section.

\subsection{Formulation of the QP Problem}
\label{sect:QP_MPC}

Collect in vectors 
\begin{eqnarray*}
\dot X_z^k &=& (\dot x_z^k \>\> \ldots \>\> \dot x_z^{k+C-1})^T\\
\dot Y_z^k &=& (\dot y_z^k \>\> \ldots \>\> \dot y_z^{k+C-1})^T\\
X_f^k &=& (x_f^{1} \>\> \ldots \>\> x_f^{F'})^T\\
Y_f^k &=& (y_f^{1} \>\> \ldots \>\> y_f^{F'})^T
\end{eqnarray*}
all the MPC decision variables. 

At this point, the QP problem can be formulated as:

\medskip

\begin{braced}
\[
\min_{\begin{array}{c} \scriptstyle \dot X_z^k,\dot Y_z^k \\ \scriptstyle X_f^k,Y_f^k \end{array}}  
\|\dot X_z^{k} \|^2  + \|\dot Y_z^{k} \|^2 + 
\beta \left( \| X_f -\hat X_f \|^2  +  \| Y_f -\hat Y_f \|^2 \right)
\]
\centerline{subject to:}

\begin{itemize}

\item ZMP constraints~(\ref{eq:ZMPcon})

\item kinematic constraints~(\ref{eq:footposcon})

\item stability constraints~(\ref{eq:StabConstrSplitCaus}) for $x$ and $y$

\end{itemize}
\end{braced}

\bigskip

Note the following points.

\begin{itemize}

\item[$\bullet$] While the ZMP and kinematic constraints involve simultaneously the $x$ and $y$ coordinates, the stability constraints must be enforced separately along the sagittal and coronal axes. 

\item[$\bullet$] The actual expression of the stability constraint will depend on the chosen tail (truncated, periodic, anticipative).

\item[$\bullet$] The same expression of the stability constraint is obtained by imposing for $x$ and $y$ the corresponding terminal constraint.

\item[$\bullet$] The CoM coordinate $x_c$ only appears through $x_u$ in the stability (or terminal) constraints.

\end{itemize}

\subsection{Generic Iteration}

We now provide a sketch of the generic iteration of the IS-MPC algorithm. The input data are the sequence $(\hat X^k_f,\hat Y^k_f, \Theta^k_f)$ of candidate footsteps,  with the associated timing ${\cal T}^k_s$, as well as the high-level reference velocities used for footstep generation (these are used explicitly in the MPC if the anticipative tail is used). As initialization, one needs $x_c$, $\dot x_c$ and $x_z$ at the current sampling instant $t_k$. Depending on the available sensors, one may either use measured data (typically true for the CoM variables) or the current model prediction (often for the ZMP position). 

The IS-MPC iteration at $t_k$ goes as follows.

\begin{enumerate}

\item Solve the QP problem to obtain $\dot X_z^k,\dot Y_z^k, X_f^k,Y_f^k$.

\item From the solutions, extract $\dot x_z^k$, $\dot y_z^k$, the first control samples. 

\item Set $\dot x_z=\dot x_z^k$ in~(\ref{eq:ThirdOrder}) and integrate from $(x_c^k,\dot x_c^k,x_z^k)$ to obtain $x_c(t)$, $\dot x_c(t)$, $x_z(t)$ for $t \in [t_k,t_{k+1}]$. Compute $y_c(t)$, $\dot y_c(t)$, $y_z(t)$ similarly.

\item Define the 3D trajectory of the CoM as $\bfp^\ast_c \!=\! (x_c,y_c,h_c)$ in $[t_k,t_{k+1}]$ and return it.

\item Return also the actual footstep sequence $(X^k_f,Y^k_f, \Theta^k_f)$ with the (unmodified) timing ${\cal T}^k_s$.

\end{enumerate}

We recall that the footstep sequence is used by the swing foot trajectory generation module  for computing $\bfp^*_{\it swg}$ in $[t_k,t_{k+1}]$ (actually, only the first footstep is needed for this computation). This is then sent to the kinematic controller together with $\bfp^\ast_c$ (see Fig.~\ref{fig:BlockSchemeDoubleStage}).

\section{IS-MPC: Feasibility and Stability}
\label{sect:feasibility}

In this section we address the crucial issues of feasibility and stability of the proposed IS-MPC controller in itself, i.e., independently from the footstep generation module. We start by reporting some simulations that show how the introduction of the stability constraint is beneficial in guaranteeing that the CoM trajectory is always bounded with respect to the ZMP trajectory. A theoretical analysis of the feasibility of the generic IS-MPC iteration is then presented and used to obtain explicit conditions for recursive feasibility; simulations are used again to confirm that the choice of an appropriate tail is essential for achieving such property. Finally, we formally prove that internal stability of the CoM/ZMP dynamics is ensured provided that IS-MPC is recursively feasible. 

\subsection{Effect of the Stability Constraint}
\label{sect:stability}

We present here some MATLAB simulation results of IS-MPC for the dynamically extended LIP model, in which we have set $h_c =0.78$~m (an appropriate value for the HRP-4 humanoid robot, see Sect.~\ref{sect:Sims}).  A sequence of evenly spaced footsteps is given with a constant step duration $T_s=0.5$~s, split in $T_{ss}=0.4$~s (single support) and $T_{ds}=0.1$~s (double support). The dimensions of the ZMP admissible regions are $d_{z,x}=d_{z,y}=0.04$~m and the sampling time is $\delta=0.01$~s. For simplicity, the footstep sequence given to the MPC is not modifiable (this corresponds to $\beta$ going to infinity in the QP cost function of Sect.~\ref{sect:QP_MPC}); correspondingly, the kinematic constraints~(\ref{eq:footposcon}) are not enforced. The QP problem is solved with the {\tt quadprog} function, which uses an interior-point algorithm.

\begin{figure}[t]
\stableLong \\
\jerkLong\\
\caption{Simulation 1: Gaits generated by IS-MPC (top) and standard MPC (bottom) for $T_c = 1.5$~s. The given footstep sequence is shown in magenta. Note the larger region corresponding to the initial double support.}
\label{fig:SimLong}
\bigskip
\stableShort  \\
\jerkShort \\
\caption{Simulation 2: Gaits generated by IS-MPC (top) and standard MPC (bottom) for $T_c = 1.0$~s. Note the instability in the standard MPC solution.}
\label{fig:SimShort}
\end{figure}

We compare the performance of the proposed IS-MPC scheme with a standard MPC. In IS-MPC, we have used~(\ref{eq:SC_Per}) as stability constraint, which corresponds to choosing a periodic tail.  In the standard MPC, the stability constraint is removed and the ZMP velocity norms in the cost function are replaced with the CoM jerk norms in order to bring the CoM into play. This corresponds to entrusting the boundedness of the CoM trajectory entirely to the cost function, in the hope that minimization of the CoM jerk will penalize diverging behaviors, as done in early MPC approaches for gait generation. 

Figure~\ref{fig:SimLong} shows the performance of IS-MPC and standard MPC for $T_c = 1.5$~s, i.e., 1.5 times the gait period. Both gaits are stable, with the IS-MPC gait more aggressively using the ZMP constraints in view of its cost function that penalizes ZMP variations. 

Figure~\ref{fig:SimShort} compares the two schemes when the control horizon is reduced to $T_c = 1$~s. The standard MPC loses stability: the resulting ZMP trajectory is always feasible but the associated CoM trajectory diverges\footnote{In particular,  the divergence occurs in this case on the coronal coordinate $y_c$. However, it is also possible to find situations where divergence occurs on the sagittal coordinate $x_c$, or even on both coordinates.} with respect to it, because the control horizon is too short to allow sorting out the stable behavior via jerk minimization.  With IS-MPC, instead, boundedness of the CoM trajectory with respect to the ZMP trajectory is preserved in spite of the shorter control horizon thanks to the embedded stability constraint. The accompanying video shows an animation of the evolutions in Figs.~\ref{fig:SimLong}--\ref{fig:SimShort}.

\begin{figure}[t]
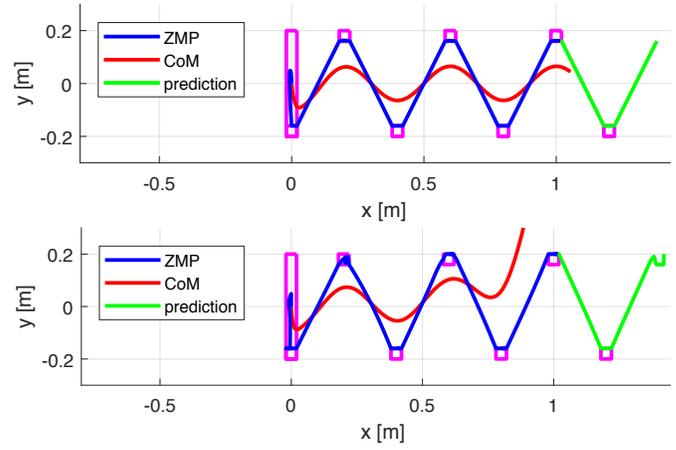

\stableLongHighCoM \\[-0.05cm]
\jerkLongHighCoM\\
\caption{Simulation 2 bis: Gaits generated by IS-MPC (top) and standard MPC (bottom) for $T_c = 1.5$~s and a higher CoM. Note the instability in the standard MPC solution.}
\label{fig:SimLongHighCoM}
\end{figure}

Another interesting situation is that of Fig.~\ref{fig:SimLongHighCoM}, in which the CoM height is increased to $h_c=1.6$~m while keeping the `long' control horizon $T_c = 1.5$~s of Simulation 1. Once again, standard MPC is unstable while IS-MPC guarantees boundedness of the CoM with respect to the ZMP. Since it is $\eta^2=g/h_c$, a similar situation can be met when $g$ is decreased, as in gait generation for low-gravity environments (e.g., the moon).

\begin{figure}[t]
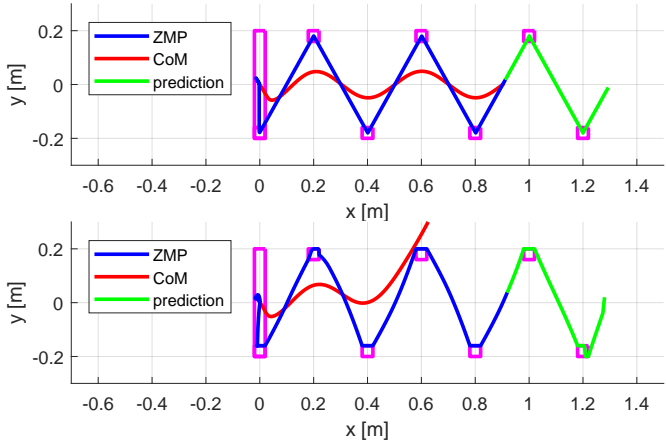

\stableSimKeepZMP \\[-0.05cm]
\jerkSimKeepZMP\\
\caption{Simulation 2 ter: Gaits generated by IS-MPC (top) and standard MPC (bottom) for $T_c = 1.0$~s, adding in the cost function a term for keeping the ZMP close to the foot center. The standard MPC solution is still unstable.}
\label{fig:SimKeepZMP}
\end{figure}

We emhasize that the onset of instability in standard MPC cannot be avoided by adding to the cost function a term for keeping the ZMP close to the foot center. The result of this common expedient is shown in Fig.~\ref{fig:SimKeepZMP}, in which the divergence occurs even earlier than in~Fig.~\ref{fig:SimShort}, because the additional cost term has actually the effect of depenalizing the norm of the CoM jerk. Instead, IS-MPC remains stable also with this modified cost function, with the ZMP pushed well inside the constraint region.

\subsection{Feasibility Analysis}

The introduction of the stability constraint (or the corresponding terminal constraint), although beneficial in guaranteeing boundedness of the CoM trajectory, has the effect of reducing the {\em feasibility region}, i.e., the subset of the state space for which the QP problem of Sect.~\ref{sect:QP_MPC} admits a solution. In some situations this might even lead to a loss of feasibility; i.e., the system may find itself in a state where it is impossible to find a solution satisfying all the constraints.

In the following we show how to determine the feasibility region at a given time. Then we address {\em recursive feasibility}: this property holds if, starting from a feasible state, the MPC scheme always brings the system to a state which is still feasible. In particular, we will prove that one can achieve recursive feasibility by using the preview information conveyed by the sequence of candidate footsteps.

\medskip

\subsubsection{Feasibility Regions}

To focus on the feasibility issue, consider the case of given footsteps ($\beta\to \infty$ in the QP cost function) with fixed orientation. 
Thanks to the latter assumption, and to the use of a moving ZMP constraint in double support (Fig.~\ref{fig:DoubleSupport}), the QP problem separates in two decoupled problems, one for the $x$ and one for the $y$ ZMP coordinate.  Let us focus on the $x$ coordinate henceforth, with the understanding that every development is also valid for the $y$ coordinate. The general coupled case can be treated by using an appropriate coordinates change.

Consider the $k$-th step of the IS-MPC algorithm. The QP problem is feasible at $t_k$ if there exists a ZMP trajectory $x_z(t)$ that satisfies both the ZMP constraint for $t\in[t_k,t_{k+C}]$ 
\begin{equation}
x_z^m(t) \leq x_z(t) \leq x_z^M(t),
\label{eq:ZMPConstr_ContTime}
\end{equation}
and the stability constraint
\begin{equation}
\eta\int_{t_k}^{t_{k+C}} \!\!\! e^{-\eta(\tau-t_k)}x_z (\tau) d\tau = x_u^k - \eta\int_{t_{k+C}}^\infty \!\!\! e^{-\eta(\tau-t_k)}\tilde x_z (\tau) d\tau,
\label{eq:StabConstr_ContTime}
\end{equation}
where:

\begin{itemize}

\item $x_z^m(t)$ and $x_z^M(t)$ are respectively the lower and upper bound of the ZMP admissible region at time $t$, as derived from~(\ref{eq:ZMPcon});

\item $\tilde x_z$ is the ZMP position\footnote{In the rest of this section, we will for simplicity use the term `tail' for both the ZMP velocity and the corresponding position.} corresponding (through integration) to the chosen velocity tail;

\item both the ZMP and the stability constraint have been expressed in continuous time for later convenience (in particular, eq.~(\ref{eq:StabConstr_ContTime}) is obtained from~(\ref{eq:bcConstr}) by splitting the integral in two and plugging the tail in the second integral);

\item the kinematic constraints~(\ref{eq:footposcon}) are not enforced since footsteps are given.
\end{itemize}

\medskip

\begin{proposition}
At time $t_k$, IS-MPC is feasible if and only if
\begin{equation}
x_u^{k,m} \leq x_u^k \leq x_u^{k,M}
\label{eq:feasibilityRegion}
\end{equation}
where 
\begin{eqnarray*}
x_u^{k,m} &\!\!\!=\!\!\!& 
\eta\int_{t_k}^{t_{k+C}} e^{-\eta(\tau-t_k)}x_z^m d\tau + 
\eta\int_{t_{k+C}}^\infty e^{-\eta(\tau-t_k)}\tilde x_z d\tau
\\
x_u^{k,M} &\!\!\!=\!\!\!& 
\eta\int_{t_k}^{t_{k+C}} e^{-\eta(\tau-t_k)}x_z^M d\tau + 
\eta\int_{t_{k+C}}^\infty e^{-\eta(\tau-t_k)}\tilde x_z d\tau.
\end{eqnarray*}
\label{prop:Fea}
\end{proposition}

\medskip

{\em Proof}.
To show the necessity of~(\ref{eq:feasibilityRegion}), multiply each side of the ZMP constraint (\ref{eq:ZMPConstr_ContTime}) by $e^{-\eta(t-t_k)}$ and integrate over time from $t_k$ to $t_{k+C}$. Adding to all sides the integral term in the right-hand side of~(\ref{eq:StabConstr_ContTime}), the middle side becomes exactly $x_u^k$, while the left- and right-hand sides become $x_u^{k,m}$ and $x_u^{k,M}$ as defined in the thesis.

The sufficiency can be proven by showing that if~(\ref{eq:feasibilityRegion}) holds then the ZMP trajectory 
\[
x_z(t) = x_z^M(t) - \frac{x_u^{k,M} - x_u^k}{1- e^{-\eta T_c}}
\]
satisfies both the ZMP constraint (\ref{eq:ZMPConstr_ContTime}) and the stability constraint~(\ref{eq:StabConstr_ContTime}).
\hfill\bull
\medskip

The interpretation of~(\ref{eq:feasibilityRegion}) is the following: it is the admissible range for $x_u$ at time $t_k$ to guarantee solvability of the QP problem associated to the current iteration of IS-MPC. Since $x_u$ is related to the state variables of the prediction model through~(\ref{eq:xutransf}), eq.~(\ref{eq:feasibilityRegion}) actually identifies the feasibility region in state space.

Note that
\begin{eqnarray}
x_u^{k,M} - x_u^{k,m} &=& \eta\int_{t_k}^{t_{k+C}} e^{-\eta(\tau-t_k)} (x_z^M-x_z^m) d\tau\nonumber\\ 
&=& d_{z,x} (1 - e^{-\eta T_c}),
\label{eq:FeaRegExt}
\end{eqnarray}
where we have used the fact that $x_z^M(t)-x_z^m(t)=d_{z,x}$ for all $t$, as implied by~(\ref{eq:ZMPcon}). This shows that the extension $x_u^{k,M} - x_u^{k,m}$ of the admissible range for $x_u$ depends on the dimension $d_{z,x}$ of the ZMP admissible region, and tends to become exactly $d_{z,x}$ as the control horizon $T_c$ is increased. On the other hand, the midpoint of this range depends on the tail chosen for the stability constraint~(\ref{eq:StabConstr_ContTime}), because $\eta\int_{t_{k+C}}^\infty e^{-\eta(\tau-t_k)}\tilde x_z d\tau$ acts as an offset in both the left- and right-hand sides of~(\ref{eq:feasibilityRegion}).

\begin{figure}[t]
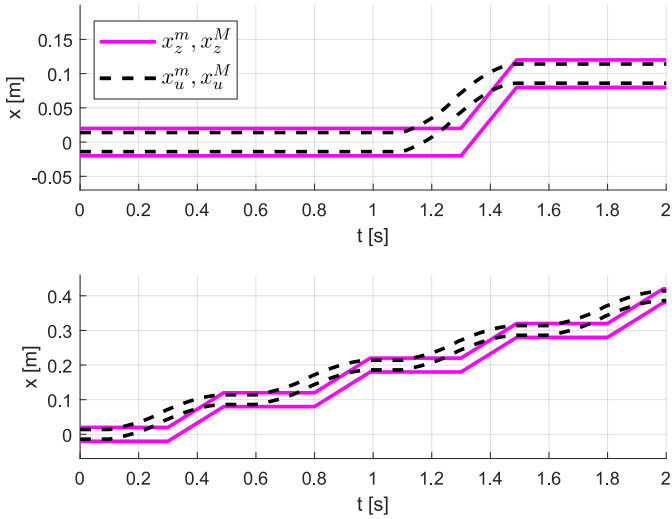

\FeasibilityRegions
\caption{Feasibility regions. Top: The robot is taking a single step. Bottom: The robot is taking a sequence of steps. The anticipative tail is used in both cases.}
\label{fig:FeasibilityRegions}
\end{figure}

Figure~\ref{fig:FeasibilityRegions} illustrates how the admissible range for $x_u$ moves over time, for the case of a single step and of a sequence of steps.
These results were obtained with $h_c=0.78$~m, $d_{z,x} = 0.04$~m and $T_c=0.5$~s. In both cases, an anticipative tail was used, with the residual part truncated; the preview horizon is $T_p=1$~s.
Note that, as expected, the extension of the range is constant and smaller than $d_{z,x}$, and that the range itself gradually shifts toward the next ZMP admissible region as a step is approached. 

\medskip

\subsubsection{Recursive Feasibility} 
\label{sect:Recursive}

We prove next that the use of an anticipative tail provides recursive feasibility under a (sufficient) condition on the preview horizon $T_p$.

\medskip

\begin{proposition}
Assume that the anticipative tail is used in the stability constraint~(\ref{eq:StabConstr_ContTime}). Then, IS-MPC is recursively feasible if the preview horizon $T_p$ is sufficiently large.
\label{prop:RecFea}
\end{proposition}

\medskip

{\em Proof}. 
To establish recursive feasibility, we must show that if the IS-MPC QP problem is feasible at $t_k$, it will be still feasible at time $t_{k+1}$. 

Let us assume that (\ref{eq:feasibilityRegion}) holds. This implies that the ZMP constraint~(\ref{eq:ZMPConstr_ContTime}) holds for $t \in [t_k, t_{k+C}]$, and that the stability constraint~(\ref{eq:StabConstr_ContTime}) is satisfied, i.e., 
\[
x_u^k = \eta\int_{t_k}^{t_{k+C}} \!\!\! e^{-\eta(\tau-t_k)}x_z d\tau + \eta\int_{t_{k+C}}^\infty \!\!\! e^{-\eta(\tau-t_k)}\tilde x_z (\tau) d\tau,
\]
with $\tilde x_z$ chosen as the anticipative tail at $t_k$. 

Using~(\ref{eq:xudyn}), the value of $x_u$ at $t_{k+1}$ is written as
\[
x_u^{k+1} = e^{\eta \delta} x_u^{k2	} - \eta \int_{t_k}^{t_{k+1}} \!\!\! e^{\eta (t_{k+1} - \tau)} x_z (\tau) d\tau.
\]
Plugging the above expression for $x_u^k$ in this equation, simplifying, and considering that $x_z(t) \leq x_z^M(t)$ for $t \in [t_k, t_{k+C}]$ we obtain
\[
x_u^{k+1} \!\!\leq\! \eta \int_{t_{k+1}}^{t_{k+C}} \!\!\!\!\!\!\! e^{\eta (t_{k+1} - \tau)} x_z^M (\tau) d\tau + \eta \int_{t_{k+C}}^\infty \!\!\!\!\!\! e^{\eta (t_{k+1} - \tau)} \tilde x_z (\tau) d\tau.
\]
According to Prop.~\ref{prop:Fea}, feasibility at $t_{k+1}$ requires\footnote{From now on, we focus only on the right-hand side of the feasibility condition for compactness. In fact, imposing the left-hand side leads to the same condition~(\ref{eq:RFSuffCond}).}
\[
x_u^{k+1} \!\!\leq\! \eta \int_{t_{k+1}}^{t_{k+C+1}} \!\!\!\!\!\!\!\!\!\!\!\!\! e^{\eta (t_{k+1} - \tau)} x_z^M (\tau) d\tau + \eta \int_{t_{k+C+1}}^\infty \!\!\!\!\!\!\!\!\!\!\!\! e^{\eta (t_{k+1} - \tau)} \tilde x'_z (\tau) d\tau,
\]
with $\tilde x'_z (\tau)$ in the second integral denoting the anticipative tail at $t_{k+1}$.
Recursive feasibility is then guaranteed if the right-hand side of the last equation is not larger than that of the penultimate. This condition can be rewritten as
\begin{align}
& \int_{t_{k+C}}^{t_{k+C+1}} \!\!\!\!\!\!\!\!\!\! e^{\eta (t_{k+1} - \tau)} \tilde x_z (\tau) d\tau
+
\int_{t_{k+P}}^{\infty}  e^{\eta (t_{k+1} - \tau)} \tilde x_z (\tau) d\tau \leq \nonumber \\
& \int_{t_{k+C}}^{t_{k+C+1}} \!\!\!\!\!\!\!\!\!\! e^{\eta (t_{k+1} - \tau)} x_z^M (\tau) d\tau
+
\int_{t_{k+P}}^{\infty} \!\!\!\!\!\!\! e^{\eta (t_{k+1} - \tau)} \tilde x'_z (\tau) d\tau, \nonumber
\end{align}
where we have used the fact that the anticipative tails at $t_k$ and $t_{k+1}$ coincide over $[t_{k+C+1},t_{k+P}]$. From this we derive the equivalent inequality
\begin{align*}
& \int_{t_{k+C}}^{t_{k+C+1}} e^{\eta (t_{k+1} - \tau)} (x_z^M (\tau) - \tilde x_z (\tau)) d\tau +\\
& \int_{t_{k+P}}^{\infty} e^{\eta (t_{k+1} - \tau)} (\tilde x'_z (\tau) - \tilde x_z (\tau)) d\tau \geq 0.
\end{align*}
At this point, exploiting the fact (see the end of Sect.~\ref{sect:PredTail}) that {\em (i)} $x_z^M (t) - \tilde x_z (t) = d_{z,x}/2$ in the preview horizon, and {\em (ii)} the residual part of the anticipative tail is truncated, a lenghty but simple calculation leads to the condition
\[
\frac{e^{-\eta (T_p-T_c)}}{\eta} (\dot{\tilde x}'_z)^{k+P} + \frac{d_{z,x}}{2} \geq 0,
\]
where $(\dot{\tilde x}'_z)^{k+P}$ is the last velocity sample in the preview horizon of the anticipative tail at $t_{k+1}$. Finally, if we denote by $v^{\rm max}_{z,x}$ the upper bound on the absolute value of $(\dot{\tilde x}'_z)^{k+P}$, we can claim that a sufficient condition for recursive feasibility is 
\begin{equation}
T_p \geq T_c + \frac{1}{\eta} \log \frac{2\, v^{\rm max}_{z,x}}{\eta \,d_{z,x}},
\label{eq:RFSuffCond}
\end{equation}
thus concluding the proof.\hfill\bull
\medskip

Note the following points.

\begin{itemize}

\item An upper bound $v^{\rm max}_{z,x}$ to be used in~(\ref{eq:RFSuffCond}) can be derived (and enforced in the tail) based on the dynamic capabilities of the specific robot or, even more directly, using the information embedded in the footstep sequence and timing. This is the same kind of reasoning that led us to postulate the existence of an upper bound $\gamma$ on $\dot x^i_z$ in~(\ref{eq:LinearZ}).

\item Equation~(\ref{eq:RFSuffCond}) shows that a longer preview horizon $T_p$ is needed to guarantee recursive feasibility for taller and/or faster robots (larger $\eta$ and/or $v^{\rm max}_{z,x}$, respectively), or for robots with more compact feet (smaller $d_{z,x}$).

\item Proposition~\ref{prop:RecFea} provides only a sufficient condition, and therefore does not exclude that recursive feasibility of IS-MPC can be achieved with a smaller preview horizon, or even with a different tail. For example, in the next subsection we will describe a case (Simulation 3) in which the periodic tail represents a sufficiently accurate conjecture and therefore recursive feasibility is achieved.

\end{itemize}

\medskip

\subsubsection{Recursive Feasibility --- Simulations} 
\label{sect:RFSim}

\begin{figure}[t]
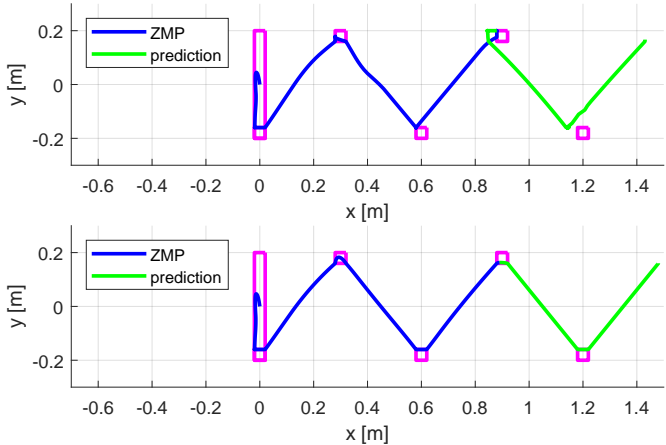

\regularTruncated\\
\regularPeriodic
\caption{Simulation 3: Gaits generated for a regular footstep sequence with different tails: truncated (top), periodic (bottom). Note the loss of feasibility when using the truncated tail.}
\label{fig:SimRegular}
\end{figure}

\begin{figure}[t]
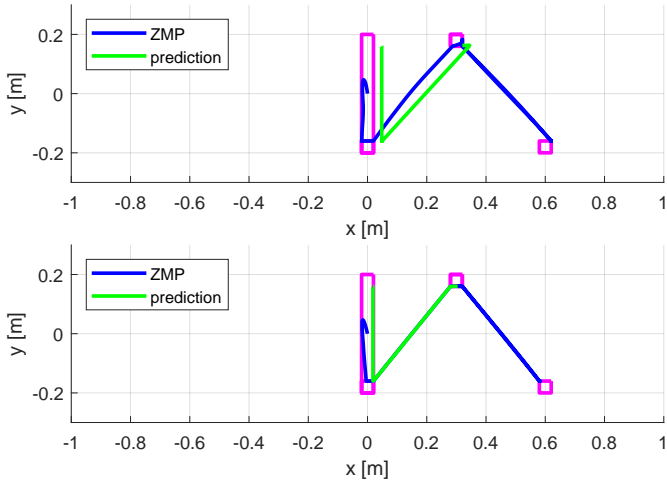

\irregularPeriodic \\
\irregularPredicted
\caption{Simulation 4: Gaits generated for an irregular footstep sequence with different tails: periodic (top), anticipative (bottom). The footstep sequence consists of two forward steps followed by two backwards steps on the same footsteps. Note the loss of feasibility when using the periodic tail.}
\label{fig:SimIrregular}
\end{figure}

We now report some comparative MATLAB simulations aimed at showing how different choices for the tail lead to different results in terms of recursive feasibility. We use the same LIP model and parameters of Sect.~\ref{sect:stability}. The MPC still operates under the assumption that the footstep sequence is given and not modifiable. The control horizon $T_c$ is 0.8~s while the preview horizon $T_p$ is 1.6~s.

Figure~\ref{fig:SimRegular} shows a comparison between IS-MPC using the truncated and periodic tail for a regular footstep sequence.  When using the truncated tail, gait generation fails because the system reaches an unfeasible state, due to the significant mismatch between the truncated tail and the persistent ZMP velocities required by the gait. Recursive feasibility is instead achieved by using the periodic tail, which 
coincides with an anticipative tail for this case.

Figure~\ref{fig:SimIrregular} refers to a situation in which the assigned footstep sequence is irregular: two forward steps are followed  by two backward steps on the same footsteps. Use of the periodic tail leads now to a loss of feasibility, as IS-MPC is wrongly conjecturing that the ZMP trajectory will keep on moving forward. The anticipative tail, which is the recommended choice for this scenario, correctly anticipates the irregularity therefore achieving recursive feasibility.

The accompanying video shows an animation of the evolutions in Figs.~\ref{fig:SimRegular}--\ref{fig:SimIrregular}.

\subsection{Recursive feasibility implies stability}
\label{sect:RFimpliesS}

In Sect.~\ref{sect:Recursive} it has been shown that recursive feasibility can be guaranteed by using the anticipative tail, provided that the preview horizon $T_p$ is sufficiently large (Proposition~\ref{prop:RecFea}). Now we prove that recursive feasibility in turn implies internal stability (i.e., boundedness of the CoM trajectory with respect to the ZMP). 

We recall a definition first. A function $f(t)$ is said to be {\em of exponential order $\alpha_0$} if~\cite{Le:61}
\[
\lim_{t \to \infty} f(t) e^{-\alpha t} = 0 \qquad \mbox{when} \quad \alpha > \alpha_0.
\]
According to this definition, any bounded or polynomial function is of exponential order $0$, whereas $e^{at}$ is of exponential order $a$. In particular, $x_z$ is of exponential order 0 in IS-MPC, because it is piecewise-linear with bounded derivative, see~(\ref{eq:LinearZ}).

\medskip

\begin{proposition}
If IS-MPC is recursively feasible, then internal stability is guaranteed. 
\label{prop:RecFeaStab}
\end{proposition}

\medskip

{\em Proof}. We establish the result by contradiction; that is, we assume that internal stability is violated, and show that this is inconsistent with IS-MPC being recursively feasible.  We focus on the dynamics along the sagittal axis $x$; an identical reasoning can be done along the coronal axis $y$.

Assume that internal stability is violated, i.e., {$x_c - x_z$} diverges. This implies that $x_u-x_z$ diverges, because {\em (i)} $x_c=(x_s+x_u)/2$ in view of~(\ref{eq:xstransf}--\ref{eq:xutransf}), and {\em (ii)} $x_s-x_z$ is bounded (in fact, its dynamics is BIBO stable and forced by $\dot x_z$, which is bounded). Since the dynamics of $x_u-x_z$ has a single eigenvalue $\eta$ and is also forced by $\dot x_z$, then $x_u-x_z$ will diverge with exponential order $\eta$. Finally, this implies that the feasibility condition~(\ref{eq:feasibilityRegion}) will be violated at a future instant of time, as the upper and lower bound in the inequality are functions of the same exponential order as $x_z$. This contradicts the assumption that IS-MPC is recursively feasible.\hfill\bull

\subsection{Wrapping up}
 
As discussed at the end of Sect.~\ref{sect:StabilityConstraint}, a causal MPC can only contain an approximate version of the stability constraint, because the tail in~(\ref{eq:StabConstrSplit}) is unknown and therefore must be conjectured. Nevertheless, Proposition~\ref{prop:RecFeaStab} states that the repeated enforcement of this constraint at each iteration of IS-MPC is effective, in the sense that {\em internal stability is achieved as long as the controller is recursively feasible}. In turn, the latter property is guaranteed if the anticipative tail is used with a $T_p$ that extends beyond $T_c$  enough to make the approximation sufficiently accurate (Proposition~\ref{prop:RecFea}, and in particular eq.~(\ref{eq:RFSuffCond})). 

At this point, the reader may wonder whether there is a requirement on the minimum control horizon $T_c$ in order for IS-MPC to work. The answer is that $T_c$ may indeed be arbitrarily small, with one caveat: as shown by eq.~(\ref{eq:FeaRegExt}), the feasibility region shrinks as $T_c$ decreases. However, once the system is initialized in this reduced region, the recursive feasibility of IS-MPC will depend only on $T_p$ through the sufficient condition~(\ref{eq:RFSuffCond}).

The possibility of decreasing $T_c$ without affecting stability is a distinct advantage of IS-MPC with respect to schemes which need sufficiently long $T_c$ to work. In fact, a shorter $T_c$ means less computation, which may be important for real-time on-board implementation on low-cost platforms, such as the NAO of our experiments.
Moreover, since the MPC needs to know the (candidate) footstep locations in the control horizon, decreasing $T_c$ means that footsteps are required over a smaller interval, making it possible to use short-term reactive planners.

\section{Simulations}
\label{sect:Sims}
  
We now report some complete gait generation results (footstep generation + IS-MPC) obtained in the V-REP simulation environment. 
The humanoid platform is HRP-4, a 34-dof, 1.5~m tall humanoid robot. We enabled dynamic simulation using the Newton Dynamics engine. 
 
The whole gait generation framework runs at 100 Hz ($\delta=0.01$~s). Footstep timing is determined using rule~(\ref{eq:TimingRule}) with $\bar L_s=0.12$~m, $\bar T_s=0.8$~s, $\bar v=0.15$~m/s as cruise parameters, and $\alpha=0.1$~m/s (as in Fig.~\ref{fig:footstepTimingRule}). 
Each generated $T_s$ is split into $T_{ss}$ (single support) and $T_{ds}$ (double support) using a 60\%-40\% distribution.
Candidate footsteps are generated as explained in Sect.~\ref{sect:Footsteps}, with $\theta_{\rm max}=\pi/8$~rad and $\ell = 0.18$~m. In the IS-MPC module, which uses a control horizon $T_c$ of 1.6~s, we have set $h_c=0.78$~m. The dimensions of the ZMP admissible region are $d_{z,x}=d_{z,y}=0.04$~m, while those of the kinematically admissible region are $d_{a,x}=0.3$~m, $d_{a,y}=0.07$~m. The weight in the QP cost function is $\beta=10^4$. The qpOASES library was used to solve the QP, here as well as in the experiments to be presented in the next section.

Figure \ref{fig:stroboStraight} shows a stroboscopic view of the first simulation (see the accompanying video for a clip). The robot is commanded a sagittal reference velocity $v_x$ of $0.1$~m/s which is then abruptly increased to $0.3$~m/s. The preview horizon is $T_p=3.2$~s and the anticipative tail is used.
The generated CoM and ZMP trajectories together with the sagittal CoM velocity are shown in Fig. \ref{fig:plotsStraight}. As expected, the higher commanded velocity is realized by increasing both step length and frequency.

In the second simulation, shown in Fig.~\ref{fig:stroboCusp} and the accompanying video, the reference velocities are aimed at producing a cusp trajectory. In particular, initially we have $v_x=0.2$~m/s and $\omega=0.2$~rad/s; after a quarter turn we change $v_x$ to $-0.2$~m/s;  after another quarter turn, $\omega$ is zeroed. As before, $T_p$ is 3.2~s and the anticipative tail is used for the stability constraint. Figure \ref{fig:plotsCusp} shows plots of the generated ZMP and CoM trajectories, together with the sagittal CoM velocity.

\begin{figure}[t]
\stroboStraight\\
\caption{Simulation 5. HRP-4 following a variable reference velocity.}	
\label{fig:stroboStraight}
\vspace{0.5cm}
\straightWalk\\[0.2cm]
\refVelocity
\caption{Simulation 5: CoM and ZMP trajectories (top), sagittal velocity (bottom).}	
\label{fig:plotsStraight}
\end{figure}

\begin{figure}[t]
\stroboCusp\\
\caption{Simulation 6: HRP-4 walking along a cusp.}
\label{fig:stroboCusp}
\vspace{0.5cm}
\CuspWalk\\[0.2cm]
\refVelocityCusp
\caption{Simulation 6: CoM and ZMP trajectories (top), sagittal velocity (bottom).}
\label{fig:plotsCusp}
\end{figure}

Video clips of the complete simulations are shown in the accompanying video.

\section{Experiments} 
\label{sect:Exps}

Experimental validation of the proposed method for gait generation was performed on two platforms, i.e., the NAO and HRP-4 humanoid robots.

NAO is a 23-dof, 58~cm tall humanoid equipped with a single-core Intel Atom running at 1.6 GHz. Our method, implemented as a custom module in the B-Human RoboCup SPL team framework~\cite{RoLaRi:15}, runs in real-time on the on-board CPU at a control frequency of 100 Hz ($\delta=0.01$~s). 
Footstep timing is determined using rule~(\ref{eq:TimingRule}) with $\bar L_s=0.075$~m, $\bar T_s=0.5$~s, $\bar v=0.15$~m/s as cruise parameters, and $\alpha=0.1$~m/s (as in Fig.~\ref{fig:footstepTimingRule}). Candidate footsteps are generated as explained in Sect.~\ref{sect:Footsteps}, with $\theta_{\rm max}=\pi/8$~rad and $\ell = 0.1$~m. In the IS-MPC module we have set $T_c=1.0$~s and $h_c=0.23$~m. The dimensions of the ZMP admissible region are $d_{z,x}=d_{z,y}=0.03$~m, while those of the kinematically admissible region are $d_{a,x}=0.1$~m, $d_{a,y}=0.05$~m. The weight in the QP cost function is $\beta=10^4$. The anticipative tail is used with a preview horizon  $T_p=2.0$~s.

The software architecture of HRP-4 requires control commands to be generated at a frequency of 200 Hz ($\delta=0.005$~s). Gait generation runs on an external laptop PC and joint motion commands are sent to the robot via Ethernet using TCP/IP. The parameters are the same of the V-REP simulations in the previous section, including $T_c=1.6$~s, with the exception of $d_{z,x}$, $d_{z,y}$ that are reduced to 0.01~m for increased safety. The anticipative tail is used in the stability constraint.

\begin{figure}[t]
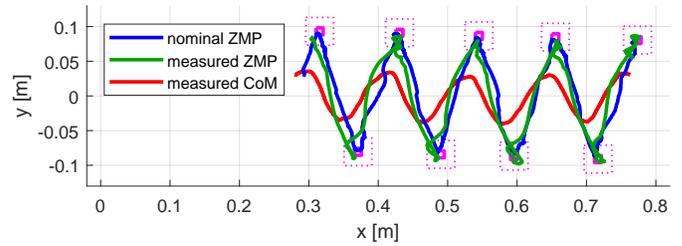

\Exp_MeasuredZMP
\caption{Nominal ZMP, measured ZMP and measured CoM along a forward gait of HRP-4. Note the restricted ZMP regions (magenta, solid) and the original ZMP regions used in the simulations (magenta, dotted).}
\label{fig:Exp_MeasuredZMP}
\end{figure}

Before presenting complete locomotion experiments, we report in Fig.~\ref{fig:Exp_MeasuredZMP} some data from a typical forward gait of HRP-4. In particular, the plot shows the nominal ZMP trajectory, as generated by IS-MPC, together with the ZMP measurements reconstructed from the force-torque sensors at the robot ankles~\cite{TaDeCoOrKh:19}. Note how the restriction of the ZMP admissible region is effective, in the sense that while the measured ZMP violates the constraints, it stays well within the original ZMP admissible region used in the simulation.

\begin{figure*}[t]
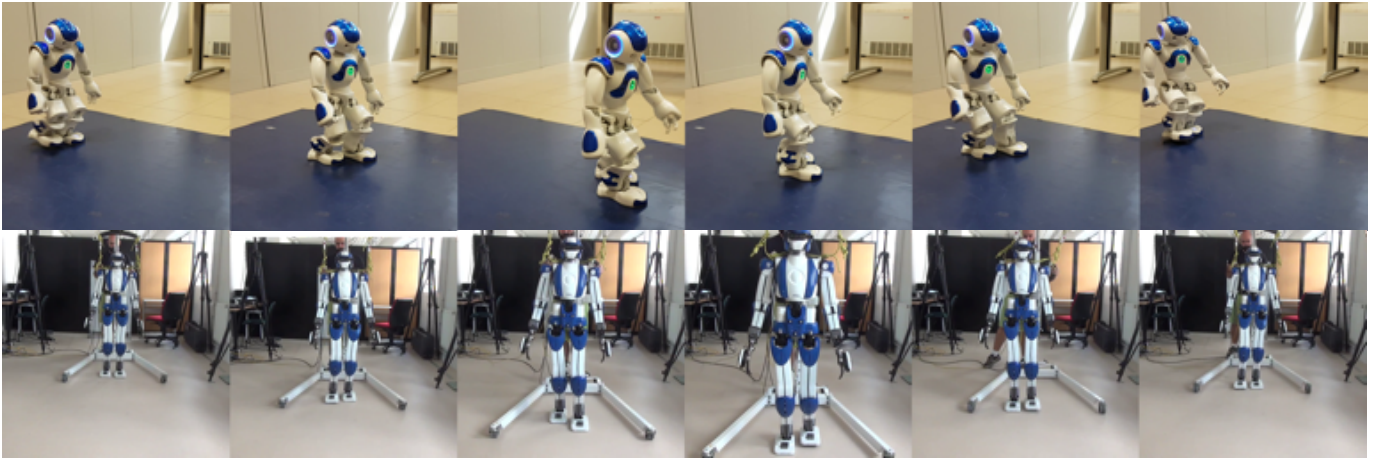

\snapshotNAOforwardbackward\\
\snapshotHRPforwardbackward
\caption{Experiments 1 and 2: NAO and HRP-4 walking forward and backward. See the accompanying video.}
\label{fig:NAO_HRP4_back_and_forth}
\end{figure*}

The accompanying video shows two successful experiments for each robot. In the first, the robots are required to perform a forward-backward motion as shown in Fig. \ref{fig:NAO_HRP4_back_and_forth}. The reference velocities are $v_x=\pm 0.15$~m/s for the NAO and $v_x=\pm 0.2$~m/s for the HRP-4. 

\begin{figure*}[t]
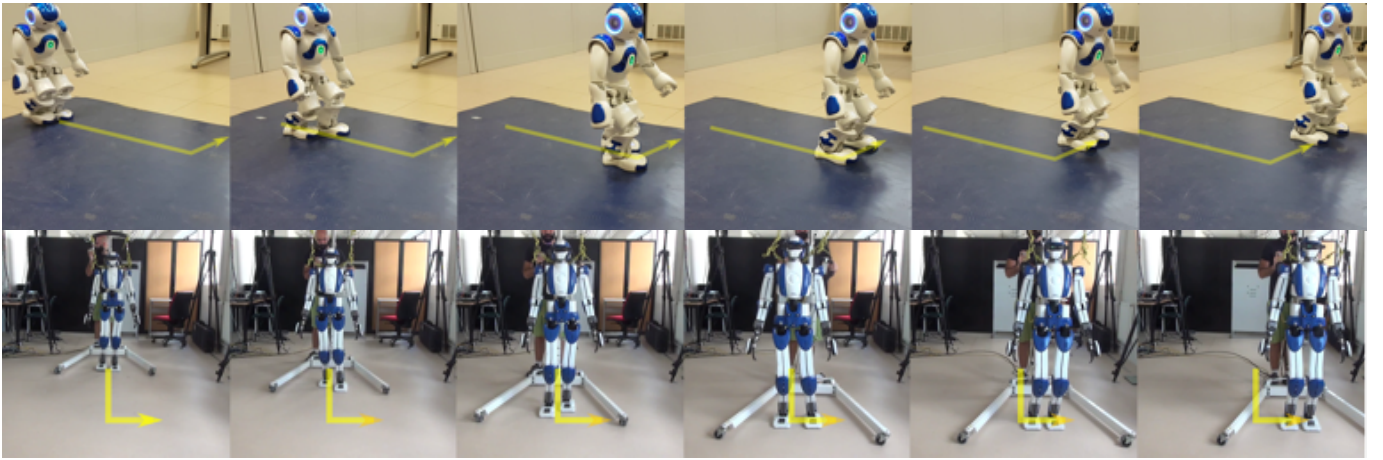

\snapshotNAOLshape\\
\snapshotHRPLshape
\caption{Experiments 3 and 4: NAO and HRP-4 walking along an L. See the accompanying video.} 
\label{fig:NAO_HRP4_L_shape}
\end{figure*}

In the second experiment, which is shown in Fig. \ref{fig:NAO_HRP4_L_shape}, the robots are given reference velocities aimed at performing an L-shaped motion. In particular, we have $v_x=0.15$~m/s followed by $v_y=0.05$~m/s for the NAO, and $v_x=0.2$~m/s followed by $v_y=0.2$ for the HRP-4. 

\section{Conclusions} 
\label{sect:Conclusions}

We have presented a complete MPC framework (IS-MPC) for generating intrinsically stable humanoid gaits that realize high-level cartesian velocity commands. We have discussed various versions of the newly introduced stability constraint, which may be used depending on the available quantity of preview information on the reference motion. It has also been shown how the different stability constraints can be interpreted as terminal constraints, some of which are new in the literature. 

A detailed study of the feasibility of the generic MPC iteration has been developed and used to derive conditions under which recursive feasibility can be guaranteed. Comparative simulations have been presented to illustrate the effect of the different tails on the resulting gait, and have confirmed that incorporating preview information in the tail is essential to preserve feasibility. Finally, it has been shown that recursive feasibility of IS-MPC implies internal stability of the CoM/ZMP dynamics.

Experimental results obtained with an on-board NAO implementation have proved that the proposed algorithm is viable even in the presence of limited computing capabilities. Additional successful experiments were carried out on the full-sized humanoid HRP-4.

The advantages of IS-MPC can be summarized as follows:

\begin{itemize} 

\item It includes an explicit stability constraint which, through the choice of the tail, can be declined on the basis of the preview information so as to accommodate different gaits to be executed. 

\item It is guaranteed to be recursively feasible if the anticipative tail is used and the preview horizon is sufficiently long (Proposition~\ref{prop:RecFea}). This clarifies the role and the amount of the required preview information, in contrast with most literature where such an analysis is missing.

\item It is the first MPC-based gait generation with an explicit proof of internal stability, which is shown to be a direct consequence of recursive feasibility (Proposition~\ref{prop:RecFeaStab}).

\item It is general enough to be applicable to different humanoids (such as NAO and HRP-4) without significant adaptation. 

\end{itemize}

We are currently working on several extensions of the proposed approach, such as:

\begin{itemize}

\item developing a robust version of the proposed IS-MPC scheme that can withstand unmodeled dynamics and disturbances~\cite{SmScMoLaOr:19};

\item extending our approach to the 2.5D case (piecewise-horizontal ground, such as stairs or flat step stones), for which we have presented a preliminary version of IS-MPC in~\cite{ZaScLaOr:18} and a footstep planner in~\cite{FeScLaOr:19};

\item investigating the use of learning techniques in conjunction with MPC in order to improve performance.

\end{itemize}

\section*{Acknowledgments}

The authors would like to thank Dr. Abderrahmane Kheddar of CNRS for hosting Daniele De Simone at LIRMM in Montpellier and allowing him to perform experiments on the HRP-4 humanoid robot.

\setcounter{equation}{0}
\renewcommand\theequation{A.\arabic{equation}}

\section*{Appendix}
\label{sect:Appendix}

We collect here some useful properties used in the proofs of the various propositions. For compactness, we use the following notation
\begin{equation}
\eta \int_{t_k}^\infty e^{-\eta(\tau-t_k)} x_z(\tau) d\tau = x_u^\ast(t_k; x_z(t)).
\label{eq:InitState}
\end{equation}

\begin{propertynew}
Linearity in $x_z(t)$:
\[
x_u^\ast(t_k; a x_z^a(t) + b x_z^b(t)) = a x_u^\ast(t_k; x_z^a(t)) + b x_u^\ast(t_k; x_z^b(t)).
\]
\end{propertynew}

\medskip

\begin{propertynew}
If $x_z(t) = \delta_{-1}(t-t_k)$, we get
\[
x_u^\ast(t_k; \delta_{-1}(t-t_k)) = 1.
\]
\end{propertynew}

\medskip

\begin{propertynew}
If $x_z(t) = \rho(t - t_k)$,  we get
\[
x_u^\ast(t_k;  \rho(t-t_k)) = 1/\eta.
\]
\end{propertynew}

\medskip

Properties 1-3 are easily derived by explicit computation of the integral in (\ref{eq:InitState}).

\medskip

\medskip

\begin{propertynew}
If $x_z(t) = 0$ for $t< t_k$, we get
\begin{equation*}
x_u^\ast(t_k; x_z(t-T)) = e^{-\eta T}x_u^\ast(t_k; x_z(t)), \quad T \ge 0.
\end{equation*}
\end{propertynew}

\medskip

{\em Proof}.
\begin{align*}
x_u^\ast(t_k; x_z(t-T)) &= \eta \int_{t_k}^\infty e^{-\eta (\tau-t_k)}x_z(\tau-T)d\tau \\
&= \eta \int_{t_{k}-T}^\infty e^{-\eta (\theta-t_k+T)}x_z(\theta)d\theta  \\
&= \eta e^{-\eta T}\int_{t_{k}-T}^\infty e^{-\eta (\theta-t_k)}x_z(\theta)d\theta  \\
&= e^{-\eta T} \eta \int_{t_{k}}^\infty e^{-\eta (\theta-t_k)}x_z(\theta)d\theta  \\
&= e^{-\eta T}x_u^\ast(t_k; x_z(t)).\qquad\qquad\qquad\qquad\hfill\bull
\end{align*}

\medskip

Property 4 ({\em time shifting)} shows how the stability condition for the time-shifted function $x_z(t-T)$ can be written in terms of the stability condition for the original function $x_z(t)$.

\input{TRO_final_ArXiv.bbl}
%

\begin{IEEEbiography}
[{\includegraphics[width=1in,height=1.25in,clip,keepaspectratio]{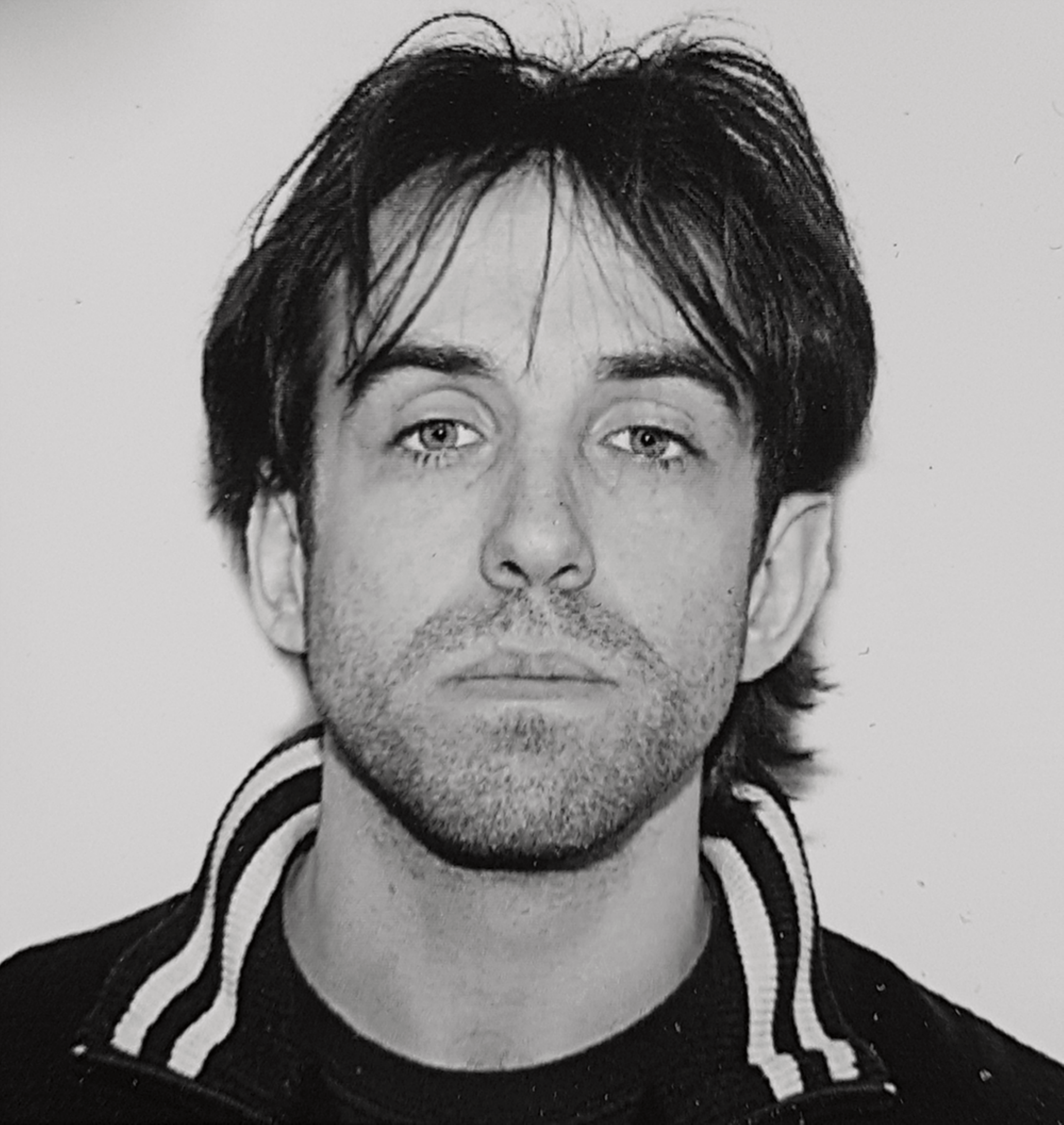}}]{Nicola Scianca} received the bachelor degree in Mechanical Engineering and the master degree in Systems Engineering, respectively in 2010 and 2014,  from Sapienza University of Rome, Italy. He is currently a Ph.D. candidate in  Control Engineering at the same university. During 2019 he was a Visiting Student at the Model Predictive Control Lab at the University of California at Berkeley. His main research interest is the use of Model Predictive Control for humanoid robots.
\end{IEEEbiography}

\begin{IEEEbiography}
[{\includegraphics[width=1in,height=1.25in,clip,keepaspectratio]{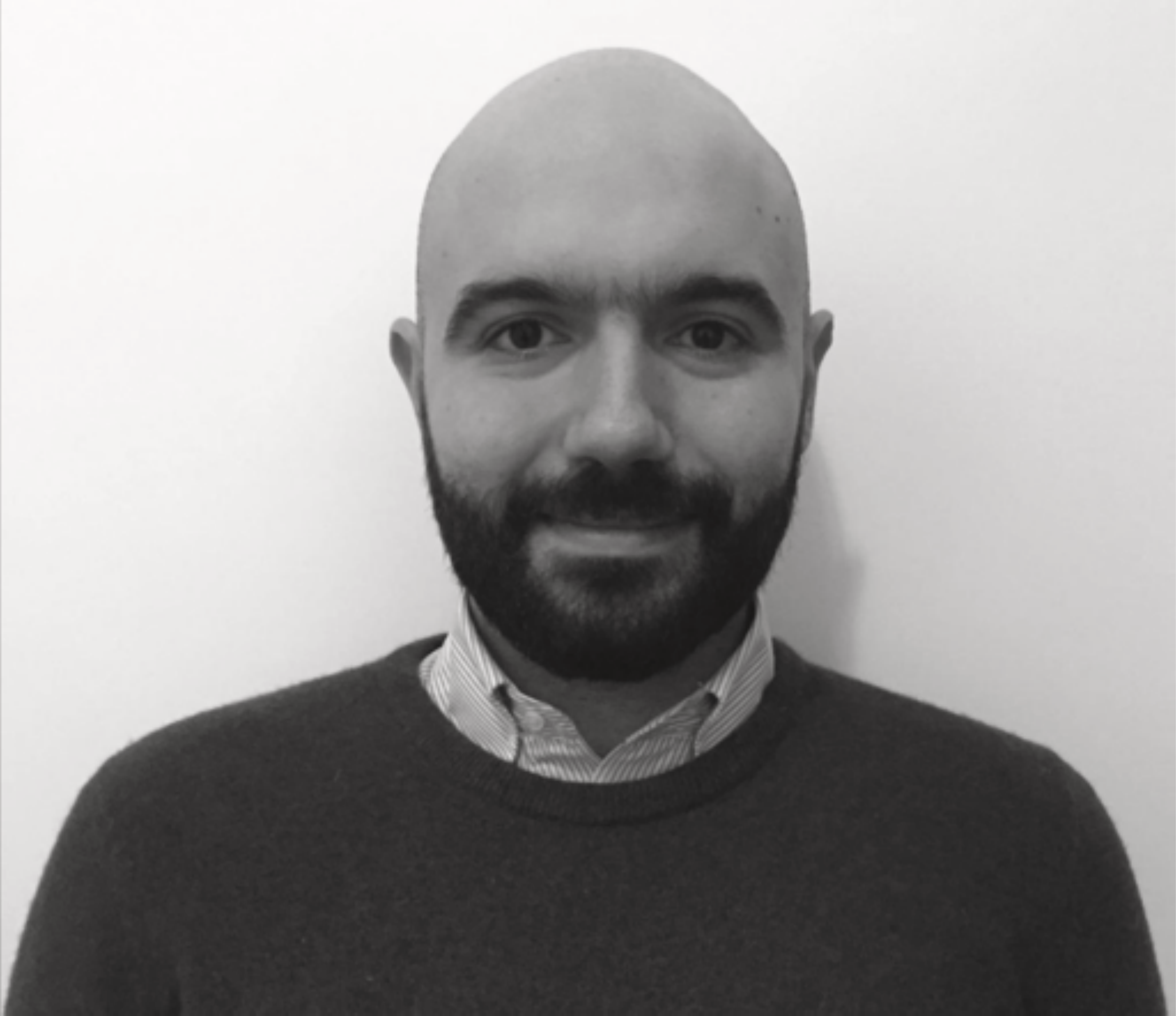}}]{Daniele De Simone} received the master degree in Artificial Intelligence and Robotics and the Ph.D. degree in Control Engineering in 2015 and 2019, respectively, from Sapienza University of Rome, Italy.  In 2018 he was a Visiting Student at the Laboratoire d'Informatique, de Robotique et de Micro\'electronique de Montpellier (LIRMM) in France. His research focuses on motion planning techniques and reactive behaviors for collision avoidance for humanoid robots.
\end{IEEEbiography}

\begin{IEEEbiography}
[{\includegraphics[width=1in,height=1.25in,clip,keepaspectratio]{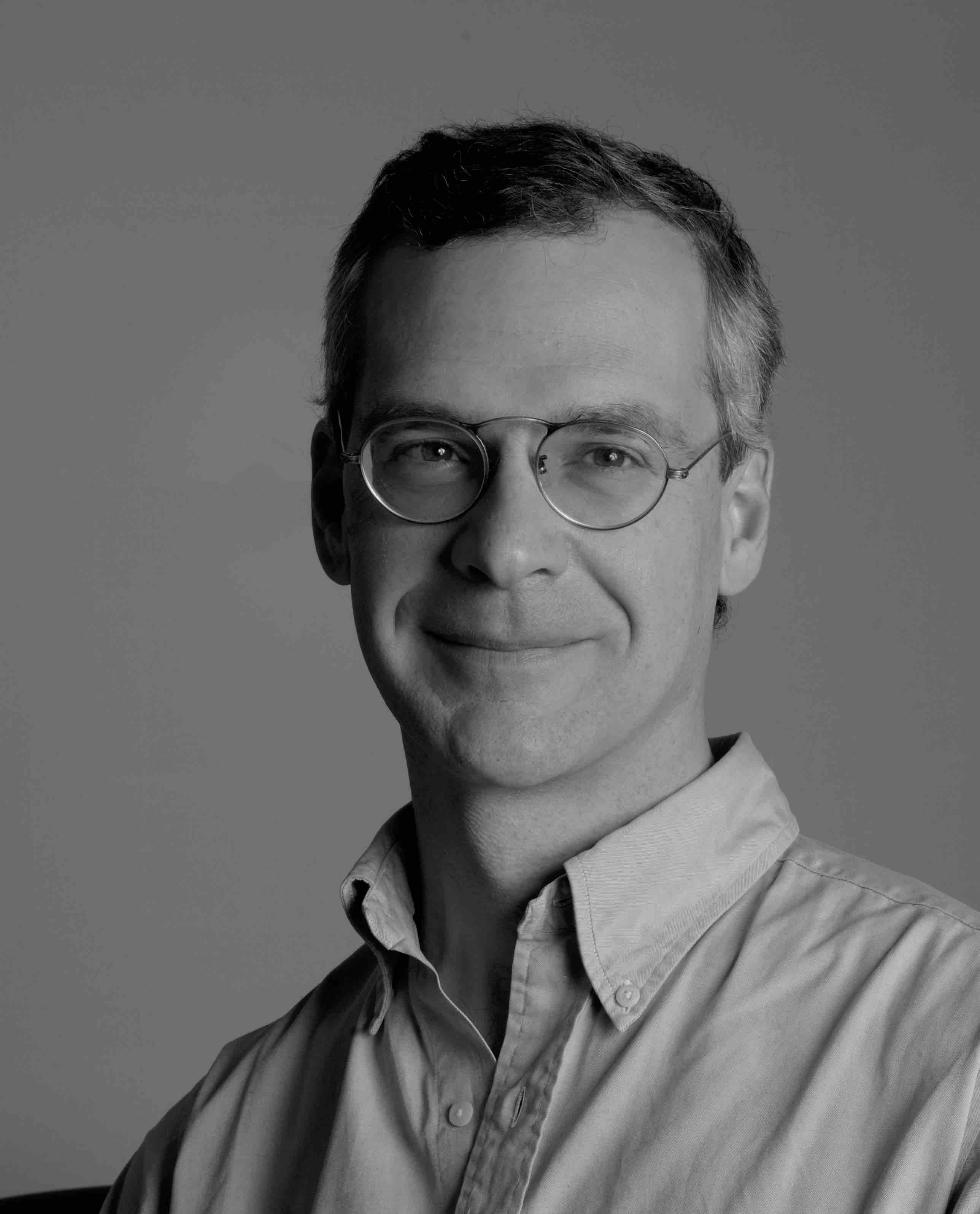}}]{Leonardo Lanari} received his Ph.D. degree in Control Engineering in 1992 from Sapienza University of Rome, Italy. He is currently with the Department of Computer, Control and Management Engineering (DIAG) of the same university as an Associate Professor in automatic control. His research interests are in the general area of control of robotic systems, with an emphasis on humanoid control and robots with elastic joints and links. 
\end{IEEEbiography}

\begin{IEEEbiography}
[{\includegraphics[width=1in,height=1.25in,clip,keepaspectratio]{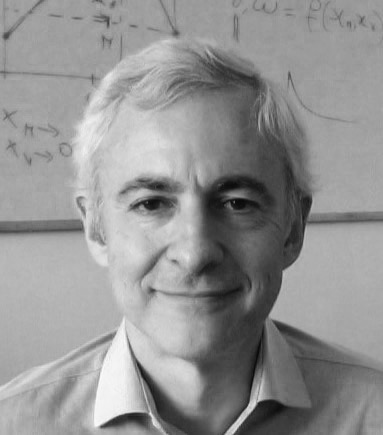}}]{Giuseppe Oriolo} (S'89-M'92-SM'02-F'16) received his Ph.D. degree in Control Engineering in 1992 from Sapienza University of Rome, Italy. He is currently with the Department of Computer, Control and Management Engineering (DIAG) of the same university, where he is a Full Professor of automatic control and robotics and the director of the DIAG Robotics Lab. His research interests are in the general area of planning and control of robotic systems. Prof. Oriolo has been an Associate Editor of the IEEE Transactions on Robotics and Automation from 2001 to 2005 and an Editor of the IEEE Transactions on Robotics from 2009 to 2013. He is a Fellow of the IEEE.
\end{IEEEbiography}

\vfill

\end{document}

%% file: mathsym.tex
\def\bfp{{\mbox{\boldmath $p$}}}

%% file: TRO_final_ArXiv.bbl
\begin{thebibliography}{10}
\providecommand{\url}[1]{#1}
\csname url@rmstyle\endcsname
\providecommand{\newblock}{\relax}
\providecommand{\bibinfo}[2]{#2}
\providecommand\BIBentrySTDinterwordspacing{\spaceskip=0pt\relax}
\providecommand\BIBentryALTinterwordstretchfactor{4}
\providecommand\BIBentryALTinterwordspacing{\spaceskip=\fontdimen2\font plus
\BIBentryALTinterwordstretchfactor\fontdimen3\font minus
  \fontdimen4\font\relax}
\providecommand\BIBforeignlanguage[2]{{%
\expandafter\ifx\csname l@#1\endcsname\relax
\typeout{** WARNING: IEEEtran.bst: No hyphenation pattern has been}%
\typeout{** loaded for the language `#1'. Using the pattern for}%
\typeout{** the default language instead.}%
\else
\language=\csname l@#1\endcsname
\fi
#2}}

\bibitem{KaHiHaYo:14}
S.~Kajita, H.~Hirukawa, K.~Harada, and K.~Yokoi, \emph{Introduction to Humanoid
  Robotics}.\hskip 1em plus 0.5em minus 0.4em\relax Springer Publishing Company
  Inc., 2014.

\bibitem{HaKaKaHi:06}
K.~Harada, S.~Kajita, K.~Kaneko, and H.~Hirukawa, ``An analytical method for
  real-time gait planning for humanoid robots,'' \emph{International Journal of
  Humanoid Robotics}, vol.~03, no.~01, pp. 1--19, 2006.

\bibitem{MoHaKaKaKaFuNaHi:06}
M.~Morisawa, K.~Harada, S.~Kajita, K.~Kaneko, F.~Kanehiro, K.~Fujiwara,
  S.~Nakaoka, and H.~Hirukawa, ``A biped pattern generation allowing immediate
  modification of foot placement in real-time,'' in \emph{6th IEEE-RAS Int.
  Conf. on Humanoid Robots}, 2006, pp. 581--586.

\bibitem{BuLoBaUlPf:07}
T.~Buschmann, S.~Lohmeier, M.~Bachmayer, H.~Ulbrich, and F.~Pfeiffer, ``A
  collocation method for real-time walking pattern generation,'' in \emph{7th
  IEEE-RAS Int. Conf. on Humanoid Robots}, 2007, pp. 1--6.

\bibitem{KaKaKaFuHaYoHi:03}
S.~Kajita, F.~Kanehiro, K.~Kaneko, K.~Fujiwara, K.~Harada, K.~Yokoi, and
  H.~Hirukawa, ``Biped walking pattern generation by using preview control of
  zero-moment point,'' in \emph{2003 {IEEE} Int. Conf. on Robotics and
  Automation}, 2003, pp. 1620--1626.

\bibitem{Wi:06}
P.-B. Wieber, ``Trajectory free linear model predictive control for stable
  walking in the presence of strong perturbations,'' in \emph{6th IEEE-RAS Int.
  Conf. on Humanoid Robots}, 2006, pp. 137--142.

\bibitem{HeDiWiDiMoDi:10}
A.~Herdt, H.~Diedam, P.-B. Wieber, D.~Dimitrov, K.~Mombaur, and M.~Diehl,
  ``Online walking motion generation with automatic footstep placement,''
  \emph{Advanced Robotics}, vol.~24, no. 5-6, pp. 719--737, 2010.

\bibitem{AlHeMa:13}
J.~Alcaraz-Jim{\'e}nez, D.~Herrero-P{\'e}rez, and H.~Mart{\'i}nez-Barber{\'a},
  ``Robust feedback control of {ZMP}-based gait for the humanoid robot {N}ao,''
  \emph{The International Journal of Robotics Research}, vol.~32, no. 9-10, pp.
  1074--1088, 2013.

\bibitem{FaPoAtIj:14}
S.~Faraji, S.~Pouya, C.~G. Atkeson, and A.~J. Ijspeert, ``Versatile and robust
  3{D} walking with a simulated humanoid robot ({A}tlas): A model predictive
  control approach,'' in \emph{2014 IEEE Int.\ Conf.\ on Robotics and
  Automation}, 2014, pp. 1943--1950.

\bibitem{GrLe:16}
R.~J. Griffin and A.~Leonessa, ``Model predictive control for dynamic footstep
  adjustment using the divergent component of motion,'' in \emph{2016 IEEE
  Int.\ Conf.\ on Robotics and Automation}, 2016, pp. 1763--1768.

\bibitem{FeXiAtKi:16}
S.~Feng, X.~Xinjilefu, C.~G. Atkeson, and J.~Kim, ``Robust dynamic walking
  using online foot step optimization,'' in \emph{2016 IEEE/RSJ Int.\ Conf.\ on
  Intelligent Robots and Systems}, 2016, pp. 5373--5378.

\bibitem{NaKuStKiMoSo:17}
M.~Naveau, M.~Kudruss, O.~Stasse, C.~Kirches, K.~Mombaur, and P.~Sou{\`e}res,
  ``A reactive walking pattern generator based on nonlinear model predictive
  control,'' \emph{IEEE Robotics and Automation Letters}, vol.~2, no.~1, pp.
  10--17, 2017.

\bibitem{CaKh:17}
S.~Caron and A.~Kheddar, ``Dynamic walking over rough terrains by nonlinear
  predictive control of the floating-base inverted pendulum,'' in \emph{2017
  IEEE/RSJ Int.\ Conf.\ on Intelligent Robots and Systems}, 2017, pp.
  5017--5024.

\bibitem{WiTeKu:16}
P.-B. Wieber, R.~Tedrake, and S.~Kuindersma, ``Modeling and control of legged
  robots,'' in \emph{Handbook of Robotics}.\hskip 1em plus 0.5em minus
  0.4em\relax Springer, 2016, pp. 1203--1234.

\bibitem{Wi:08}
P.~B. Wieber, ``Viability and predictive control for safe locomotion,'' in
  \emph{2008 IEEE/RSJ Int.\ Conf.\ on Intelligent Robots and Systems}, 2008,
  pp. 1103--1108.

\bibitem{MaRaRaSc:00}
D.~Mayne, J.~Rawlings, C.~Rao, and P.~Scokaert, ``Constrained model predictive
  control: Stability and optimality,'' \emph{Automatica}, vol.~36, no.~6, pp.
  789--814, 2000.

\bibitem{HeOtRo:14}
B.~Henze, C.~Ott, and M.~A. Roa, ``Posture and balance control for humanoid
  robots in multi-contact scenarios based on model predictive control,'' in
  \emph{2014 IEEE/RSJ Int.\ Conf.\ on Intelligent Robots and Systems}, 2014,
  pp. 3253--3258.

\bibitem{ShDiWi:14}
A.~Sherikov, D.~Dimitrov, and P.~B. Wieber, ``Whole body motion controller with
  long-term balance constraints,'' in \emph{14th IEEE-RAS Int. Conf. on
  Humanoid Robots}, 2014, pp. 444--450.

\bibitem{KoDeReGoPr:12}
T.~Koolen, T.~de~Boer, J.~Rebula, A.~Goswami, and J.~Pratt,
  ``Capturability-based analysis and control of legged locomotion, part 1:
  Theory and application to three simple gait models,'' \emph{Int. J. of
  Robotics Research}, vol.~31, no.~9, pp. 1094--1113, 2012.

\bibitem{SuYa:17}
T.~{Sugihara} and T.~{Yamamoto}, ``Foot-guided agile control of a biped robot
  through {ZMP} manipulation,'' in \emph{2017 IEEE/RSJ Int. Conf. on
  Intelligent Robots and Systems}, 2017, pp. 4546--4551.

\bibitem{CaBuMa:17}
J.~Carpentier, R.~Budhiraja, and N.~Mansard, ``Learning feasibility constraints
  for multi-contact locomotion of legged robots,'' in \emph{2017 Robotics:
  Science and Systems}, 2017.

\bibitem{TaMaYo:09}
T.~Takenaka, T.~Matsumoto, and T.~Yoshiike, ``Real time motion generation and
  control for biped robot - 1st report: Walking gait pattern generation,'' in
  \emph{2009 Int. Conf. on Intelligent Robots and Systems}, 2009, pp.
  1084--1091.

\bibitem{KaKaKuTaShiTaYo:17}
T.~Kamioka, H.~Kaneko, M.~Kuroda, C.~Tanaka, S.~Shirokura, M.~Takeda, and
  T.~Yoshiike, ``Dynamic gait transition between walking, running and hopping
  for push recovery,'' in \emph{17th IEEE-RAS Int. Conf. on Humanoid Robots},
  2017, pp. 1--8.

\bibitem{EnOtAl:15}
J.~Englsberger, C.~Ott, and A.~Albu-Sch\"affer, ``Three-dimensional bipedal
  walking control based on divergent component of motion,'' \emph{IEEE
  Transactions on Robotics}, vol.~31, no.~2, pp. 355--368, 2015.

\bibitem{CaEsLaMa:20}
S.~Caron, A.~Escande, L.~Lanari, and B.~Mallein, ``Capturability-based pattern
  generation for walking with variable height,'' \emph{{\rm to appear in} IEEE
  Trans.\ on Robotics}.

\bibitem{KrEnWiOt:12}
M.~Krause, J.~Englsberger, P.-B. Wieber, and C.~Ott, ``Stabilization of the
  capture point dynamics for bipedal walking based on model predictive
  control,'' in \emph{10th IFAC Symp.\ on Robot Control}, 2012, pp. 165--171.

\bibitem{Ke:01}
E.~C. Kerrigan, \emph{Robust Constraint Satisfaction: Invariant Sets and
  Predictive Control}.\hskip 1em plus 0.5em minus 0.4em\relax Ph.D.
  dissertation, University of Cambridge, 2001.

\bibitem{CiWiFr:16}
M.~Ciocca, P.-B. Wieber, and T.~Fraichard, ``Strong recursive feasibility in
  model predictive control of biped walking,'' in \emph{17th IEEE-RAS Int.
  Conf. on Humanoid Robots}, 2017, pp. 730--735.

\bibitem{Sh:16}
A.~Sherikov, ``Balance preservation and task prioritization in whole body
  motion control of humanoid robots,'' Ph.D. dissertation, Grenoble Alpes,
  2016.

\bibitem{ScCoDeLaOr:16}
N.~Scianca, M.~Cognetti, D.~De~Simone, L.~Lanari, and G.~Oriolo,
  ``Intrinsically stable {MPC} for humanoid gait generation,'' in \emph{16th
  IEEE-RAS Int. Conf. on Humanoid Robots}, 2016, pp. 101--108.

\bibitem{AbScDeLaOr:17}
A.~Aboudonia, N.~Scianca, D.~De~Simone, L.~Lanari, and G.~Oriolo, ``Humanoid
  gait generation for walk-to locomotion using single-stage {MPC},'' in
  \emph{17th IEEE-RAS Int. Conf. on Humanoid Robots}, 2017, pp. 178--183.

\bibitem{PrCaDrGo:06}
J.~Pratt, J.~Carff, S.~Drakunov, and A.~Goswami, ``Capture point: A step toward
  humanoid push recovery,'' in \emph{6th IEEE-RAS Int.\ Conf.\ on Humanoid
  Robots}, 2006, pp. 200--207.

\bibitem{LaHu:15}
L.~Lanari and S.~Hutchinson, ``Inversion-based gait generation for humanoid
  robots,'' in \emph{2015 IEEE/RSJ Int. Conf. on Intelligent Robots and
  Systems}, 2015, pp. 637--642.

\bibitem{Le:61}
W.~Le{P}age, \emph{Complex variables and the {L}aplace transform for
  engineers}.\hskip 1em plus 0.5em minus 0.4em\relax McGraw Hill Book Company,
  1961.

\bibitem{RoLaRi:15}
T.~R{\"o}fer, T.~Laue, J.~Richter-Klug, M.~Sch{\"u}nemann, J.~Stiensmeier,
  A.~Stolpmann, A.~St{\"o}wing, and F.~Thielke, ``{B}-{H}uman team report and
  code release 2015,'' 2015, only available online:
  \url{http://www.b-human.de/downloads/publications/2015/CodeRelease2015.pdf}.

\bibitem{TaDeCoOrKh:19}
A.~Tanguy, D.~De~Simone, A.~I. Comport, G.~Oriolo, and A.~Kheddar,
  ``Closed-loop {MPC} with dense visual {SLAM}-stability through reactive
  stepping,'' in \emph{2019 {IEEE} Int. Conf. on Robotics and Automation},
  2019, pp. 1397--1403.

\bibitem{SmScMoLaOr:19}
F.~M. Smaldone, N.~Scianca, V.~Modugno, L.~Lanari, and G.~Oriolo, ``Gait
  generation using intrinsically stable {MPC} in the presence of persistent
  disturbances,'' in \emph{19th IEEE-RAS Int. Conf. on Humanoid Robots}, 2019,
  pp. 682--687.

\bibitem{ZaScLaOr:18}
A.~Zamparelli, N.~Scianca, L.~Lanari, and G.~Oriolo, ``Humanoid gait generation
  on uneven ground using intrinsically stable {MPC},''
  \emph{IFAC-PapersOnLine}, vol.~51, pp. 393--398, 2018.

\bibitem{FeScLaOr:19}
P.~Ferrari, N.~Scianca, L.~Lanari, and G.~Oriolo, ``An integrated motion
  planner/controller for humanoid robots on uneven ground,'' in \emph{18th
  European Control Conf.}, 2019, pp. 1598--1603.

\end{thebibliography}
